\documentclass[letterpaper, 10 pt, conference]{ieeeconf}  % Comment this line out if you need a4paper

\IEEEoverridecommandlockouts                              % This command is only needed if 
\usepackage{graphicx}
\usepackage{graphbox}
\usepackage{fixmath}
\usepackage{amsmath}
\usepackage{amssymb}

\usepackage[dvipsnames]{xcolor}
\usepackage{cleveref}
\usepackage{subcaption}
\usepackage{ifthen}
\usepackage{xspace}

\PassOptionsToPackage{hyphens}{url}
\usepackage{color}

\newcommand{\gtlocations}{\mathbold{G^\#}}
\newcommand{\gtsensedlocations}{\mathbold{X^\#}}
\newcommand{\gtsensedvalues}{\mathbold{Y^\#}}
\newcommand{\sensedlocations}{\mathbold{X}}
\newcommand{\sensedvalues}{\mathbold{Y}}
\newcommand{\location}{g}
\newcommand{\sensedlocation}{x}
\newcommand{\quantiles}{Q}
\newcommand{\quantile}{q}
\newcommand{\estimatedquantilevalues}{\tilde{V}}
\newcommand{\estimatedquantilevalue}{\tilde{v}}
\newcommand{\quantilevalues}{V}

\newcommand{\quantilespatiallocations}{\mathcal{Q}}
\newcommand{\estimatedquantilespatiallocations}{\tilde{\quantilespatiallocations}}

\newcommand{\score}{l_s}
\newcommand{\numtiles}{|\quantiles|}

\newcommand{\objectivefunction}{f}
 
\newcommand{\argmin}{\arg\!\min} 

\renewcommand{\t}[1]{{\textrm{#1}}}

\newcommand{\thirdcolumn}{.32\columnwidth}

\newcommand{\threeboxplot}{0.26\columnwidth}
\newcommand{\fourboxplot}{.45\columnwidth}

\newcommand{\boxplotheight}{3.2cm}

\newif\ifshowrev

\newcommand*{\rev}[1]{
\ifshowrev
\leavevmode\unskip {\color{red}\textrm{#1}} \unskip
\else 
\leavevmode\unskip #1 \unskip
\fi 
}

\begin{document}
%\mainmatter              % start of a contribution
%
\title{Informative Path Planning to Estimate Quantiles for Environmental Analysis}
\author{Isabel M. Rayas~Fern\'andez\textsuperscript{\textdagger,1}, Christopher E. Denniston\textsuperscript{\textdagger,1}, David A. Caron\textsuperscript{2}, Gaurav S. Sukhatme\textsuperscript{\textdaggerdbl.1}%
\thanks{\textsuperscript{\textdagger} Equal contribution. All authors are with the University of Southern California. \textsuperscript{1} Robotic Embedded Systems Lab, \textsuperscript{2} Caron Protistan Aquatic Microbial Ecology Lab \tt \{rayas,cdennist,dcaron,gaurav\}@usc.edu}\thanks{\textsuperscript{\textdaggerdbl} G.S. Sukhatme holds concurrent appointments as a Professor at USC and as an Amazon Scholar. This paper describes work performed at USC and is not associated with Amazon.}
\thanks{This work was supported in part by the Southern California Coastal Water Research Project Authority under prime funding from the California State Water Resources Control Board on agreement number 19-003-150 and in part by USDA/NIFA award 2017-67007-26154.}
\thanks{This material is based upon work supported by the National Science Foundation Graduate Research Fellowship Program under Grant No. DGE-1842487. Any opinions, findings, and conclusions or recommendations expressed in this material are those of the author(s) and do not necessarily reflect the views of the National Science Foundation.}}%

%Control displaying changes in latex
\showrevfalse

\maketitle              % typeset the title of the contribution

\begin{abstract}
Scientists interested in studying natural phenomena often take physical specimens from locations in the environment for later analysis.
These analysis locations are typically specified by expert heuristics.
Instead, we propose to choose locations for scientific analysis by using a robot to perform an informative path planning survey. The survey results in a list of locations that correspond to the quantile values of the phenomenon of interest.
We develop a robot planner using novel objective functions to improve the estimates of the quantile values over time and an approach to find locations which correspond to the quantile values.
We test our approach in four different environments using previously collected aquatic data and validate it in a field trial.
Our proposed approach to estimate quantiles has a 10.2\% mean reduction in median error when compared to a baseline approach which attempts to maximize spatial coverage. 
Additionally, when localizing these values in the environment, we see a 15.7\% mean reduction in median error when using cross-entropy with our loss function compared to a baseline.

%\keywords{}
\end{abstract}

\section{Introduction}

\begin{figure}[h!]
\centering
    \includegraphics[width=\columnwidth,height=5.5cm]{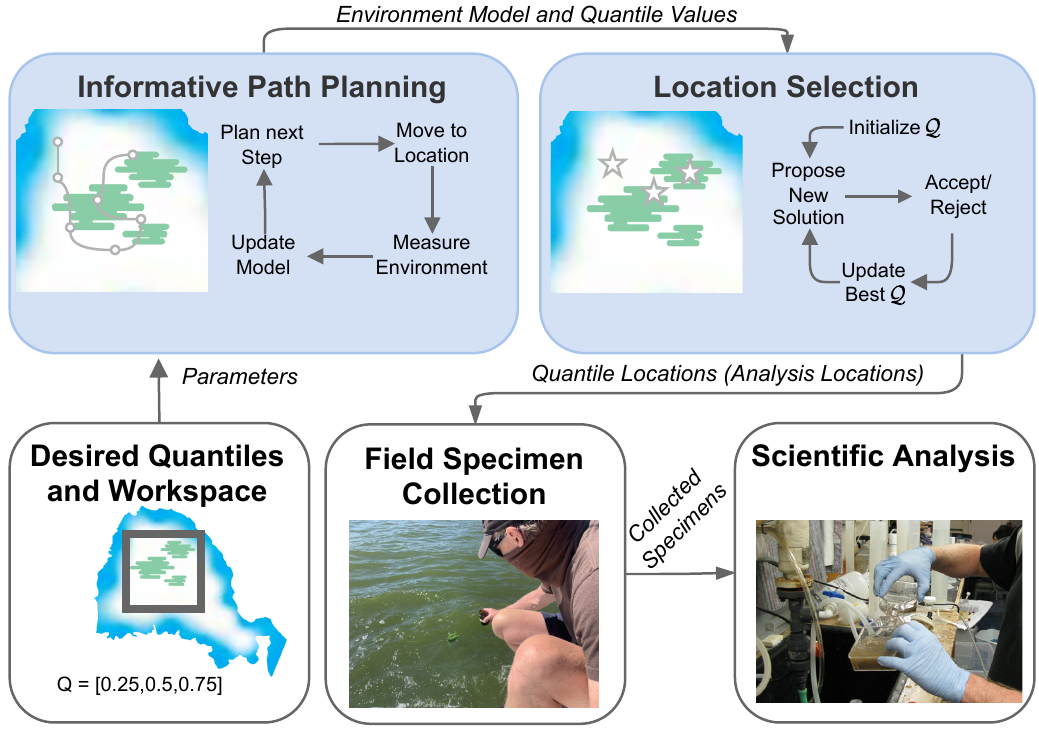}
    \caption{\textbf{Full system.}
    First, the parameters for the robotic survey are chosen, such as the area bounds and the quantiles for specimen collection.
    The robot performs informative path planning \rev{using our proposed objective functions}, creating an environment model and an estimate of the quantile values.
    \rev{The quantile locations are then selected by minimizing our proposed loss function. These quantile locations correspond to physical locations which, when measured, have the estimated quantile values.}
    After the quantile locations are chosen, humans go to them to collect field specimens which they later analyze in a laboratory setting. 
    This work focuses on the steps shown in the blue shaded boxes.
    }
    \label{fig:hero}
\vspace{-16pt}
\end{figure}

%%MOTIVATION

%Biologists take samples
In order to understand biological phenomena, scientists take a small number of physical specimens for later analyses to characterize the biological community and the contextual environmental conditions at the site.
% For example, marine biologists may be interested in using the concentration of chlorophyll present in a body of water to guide physical sampling of algal blooms, or agricultural scientists may be interested in using crop health to guide physical sampling of plants.
A marine biologist may capture water in a container or filter media at a location for later analysis, and an agricultural scientist may capture a small portion of a plant.
Scientists later analyze these captured portions of the environment in a detailed laboratory setting.
It is often expensive to collect these physical specimens as the scientist must go to the location and use a physical reagent, such as a container or filter media, which are limited in the field. 
%They don't do a good job because they use heuristics
Traditionally, expert heuristics underpin the selection of locations for scientific analysis. 
These expert heuristics generally attempt to spread out the analysis locations in the phenomena of interest in order to take specimens with differing concentrations for characterization of heterogeneity~\cite{ysi_whitepaper}.
\rev{The heuristics may make use of robotic surveys, but currently most robotic surveys use either pre-programmed paths~\cite{ysi_whitepaper}, or autonomy which seeks maxima~\cite{Marchanta} or to cover the area spatially~\cite{kemna_pilot_2018,guestrin_near-optimal_2005}, and do not directly plan for the specimen collection tasks.}
%We propose that they should choose quantiles, not necessarily evenly spaced, and the amount of which is the number of samples they can take
In contrast, we propose performing an adaptive robotic survey to find locations of interest for scientific analysis.
To specify these locations, we propose calculating the quantiles of the distribution of interest so that scientists can capture specimens at varied locations in the environment based on differing values of some phenomenon of interest to them. 
%In the case of algae bloom sampling, if you are doing a generic study and can take 10 samples maybe you want deciles, or if you only care about bad part take 3 samples at .9 95 99
For instance, if a marine biologist is interested in taking 9 physical water specimens that are spread over a range of chlorophyll concentrations, they may choose to use the deciles of the concentration. 
If only a small number of locations can be analyzed and the upper extrema values are of interest, they may choose to perform analysis at the locations of the $(0.90, 0.95, 0.99)$ quantiles.
The quantiles of interest are a flexible way to describe the objective of specimen collection that is largely invariant to the exact phenomena being measured.

%%WHAT WE DO
%Explore a region of interest
Our goal is to select locations for detailed analysis by scientists in two steps. 
We use a robot to perform an adaptive survey, and then find locations to suggest for specimen collection based on the measurements it takes.
Specifically, we \rev{first} aim to find the desired quantile values of the measurement distribution by adaptively selecting robot measurement locations that maximize an objective function designed to estimate quantile values, and then we produce suggested locations for specimen collection that are likely to contain \rev{these} values.
\rev{\Cref{fig:hero} summarizes the approach, showing how our objective function for informative path planning and our loss function for location selection fit into an overall physical specimen collection pipeline.}

% We develop our method using an informative path planning formulation
% and show that it outperforms entropy- and sequential Bayesian optimization-based objective baselines. 
Our contributions are:

\begin{itemize}
    \item A planner with custom objective functions \rev{that adaptively improve} the quality of estimated quantile values;
    \item \rev{A general loss function, which can be used with any optimizer, for selecting spatial locations for physical specimen collection} which have the values estimated for the quantiles;
    \item Quantitative evaluation of our method on point and camera sensors with previously collected aquatic datasets;
    \item A demonstration of our method on a robot in a real-world crop health estimation task.
\end{itemize}

%\section{Related Work}

\section{Background}
\textbf{Online Informative Path Planning (IPP)} consists of alternating between planning and taking an action, which usually corresponds to moving and measuring a value at the new location.
A plan described by a partial trajectory $p$ is created to maximize some objective function $\objectivefunction$ over the measured locations $\sensedlocations$ and measured values $\sensedvalues$. 
The combined plan $P$ is the concatenation of partial trajectories $p$.
The plan and act steps iterate until the cost $c(P)$ exceeds some predefined budget~\cite{Hollinger2014,denniston_icra_2020}.
% Formally, this can be described by %\cref{eq:online_informative_path_planning}.
% %\begin{equation}\label{eq:online_informative_path_planning}
% $P^* = \argmax_{P \in \Phi} f(P) ~ | ~  c(P) \leq B$,
% %\end{equation}
% where $\Phi$ is the space of full trajectories, and $P^*$ is the optimal trajectory}. 

\textbf{IPP as a Partially Observable Markov Decision Process (POMDP)} provides a formulation for planning for taking measurements.
POMDPs are a framework that can determine optimal actions when the environment is not fully observable or there is uncertainty in the environment.
\rev{Formulating IPP as a partially observable Markov decision process (POMDP) has been explored~\cite{borenstein_bayesian_2014}; 
previous works have used this for finding maxima~\cite{Marchanta} and to adapt the parameters of a rollout-based POMDP solver online to improve its efficiency~\cite{denniston_icra_2020}.
We use these works to formulate and solve the IPP problem.}

Using observations as measurements from the environment and a GP to represent the belief distribution, sequential Bayesian optimization can be formulated as a Bayesian search game~\cite{borenstein_bayesian_2014}. 
Measurements from the environment \rev{$\sensedvalues_t=GT(\sensedlocations_t)$} constitute observations which are partially observable components of the overall environment $GT$.
To adapt the Bayesian search game formulation to IPP, the belief state is augmented with the state of the robot $g_t$ and the actions the planner is allowed to take are restricted to local movements which are feasible for the robot~\cite{Marchanta}.
A complete description of the POMDP for IPP can be seen in \Cref{tab:BayesianSearchAndPOMDPs}.
We approximately solve this POMDP to select actions to take measurements for maximizing the objective function.

\begin{table}[]
\vspace{2pt}
\begin{tabular}{c|c}
\textbf{POMDP} & \textbf{Informative Path Planning}        \\ \hline
States         & Robot position $\location_t$, Underlying unknown function $GT$ \\ \hline
Actions        & Neighboring search points                     \\ \hline
Observations   & Robot position $\location_t$, Measured location(s) \rev{$\sensedlocations_t = o(\location_t)$}   \\ 
               & Measured value(s) \rev{$\sensedvalues_t = GT(\sensedlocations_t) $} \\\hline
Belief         & \rev{GP conditioned on previously measured }     \\ 
& \rev{locations  $(\sensedlocations_{0:t-1})$ and values $(\sensedvalues_{0:t-1})$} \\ \hline
Rewards        & \rev{$\objectivefunction(\sensedlocations_t)$}      
\end{tabular}
\caption{\textbf{Informative Path Planning as a POMDP.} After~\cite{borenstein_bayesian_2014,denniston_icra_2020}. }
\label{tab:BayesianSearchAndPOMDPs}
\vspace{-0.25in}
\end{table}

\textbf{Gaussian Processes (GPs)} are non-parametric models with uncertainty quantification which are widely used for IPP~\cite{Marchanta,denniston_icra_2020,kemna_pilot_2018}.
They approximate an unknown function from its known outputs by computing the similarity between points from a kernel function, in our case the squared exponential kernel~\cite{Rasmussen2006}.
GPs have been used to represent the belief distribution from observations in POMDP formulations of sequential Bayesian optimization and IPP~\cite{borenstein_bayesian_2014,Marchanta,denniston_icra_2020}.
In this work, we use a GP as the environment model to estimate the value $\mu(\sensedlocation)$ and variance $\sigma^2(\sensedlocation)$ at a specific location $\sensedlocation$.

\rev{\textbf{Objective Functions for IPP}
are used to specify the mission objective to maximize during planning. We describe some history here and show that there is a gap in the current literature for estimating quantile values.}

Often, the objective in IPP is to generate good coverage of a distributed phenomenon using an information theoretic objective such as entropy~\cite{kemna_pilot_2018} or mutual information~\cite{guestrin_near-optimal_2005}. 
Our works differs from these by instead focusing on improving estimates for specific quantiles instead of broad understanding of the spatial makeup of a distribution.

Other common objectives include finding the highest concentration location~\cite{Marchanta} or measuring near hotspots~\cite{mccammon_topological_2018}.
Sequential Bayesian optimization objectives are used to locate extreme values or areas of high concentration and have shown success in doing so in simulation and in real field scenarios~\cite{Marchanta,blanchard_informative_2021,souza_bayesian_2014}.
These specialized objective functions are typically tuned to locate a single point, often the extrema.
%, by increasing the quality of the estimate of the underlying concentration.
%Our work differs from work which seeks to find these maxima by instead finding values at different quantiles of the distribution of interest.
%
\rev{There has also been work on extending IPP missions to domains such as underwater inspection by modifying the GP and objective functions~\cite{Hollinger2013ActivePF}.
Specialized objective functions have also been developed for the task of modeling continuous and discrete variables, as well as including sensor-specific characteristics such as the increase in uncertainty due to camera distance to the subject~\cite{popovic_informative_2020}.}

\rev{Other objective functions have been proposed for level set estimation which seek to classify points into sets above and below a pre-specified concentration, or a fraction of the concentration~\cite{level_set}.
While these works have similarities with quantile estimation tasks, they require specifying a single quantile or are restricted to finding the extrema, as opposed to a set of arbitrary quantiles.}

%Adaptive sampling for quantile estimation can be compared to Bayesian optimization objectives when extreme quantiles are desired. 

%These Bayesian optimization algorithms do not have a direct method to use for locating non-extrema quantiles.
% Previous works have also investigated directly calculating the quantile values for distributions at points using specialized loss functions to extend the notion of uncertainty in Gaussian processes\rev{. However,} these do not directly transfer to our setting because they only estimate the quantiles of a specific location~\cite{boukouvalas_gaussian_2012,reich_spatiotemporal_2012} rather than of the entire distribution.

\rev{\textbf{Physical Specimen Collection} is the process of collecting portions of the environment for later analysis, which we choose using the quantiles of the distribution of the measured value in this work.}
\rev{The choice of physical specimen collection locations is driven by the spatial heterogeneity of the studied phenomena~\cite{ysi_whitepaper}.}

%The following is from Dave Caron
\rev{Spatial heterogeneity in the distribution of algal and cyanobacterial blooms in freshwater and
marine ecosystem is well known. 
Heterogeneity is often obvious as conspicuous accumulations
at down-wind and down-current locations at the surface of lakes, and vertical heterogeneity can
occur due to water column stratification, differential growth at different depths, or active vertical migration by some planktonic species~\cite{Seegers2015,Hozumi2020}. 
Such heterogeneity thwarts studies designed to quantify the spatial distribution of algal biomass, and the concentrations of algal toxins that may be produced by some algae and cyanobacteria.
Nonetheless, accurately characterizing this heterogeneity is fundamental for investigating average conditions (indicated by median quantiles) for investigating trends across many lakes or large geographical areas~\cite{Ho2019}. %Dave had Mantzouki2018 cited here but it's almost a half page long citation
Assessing heterogeneity is also essential for assessing the worst case scenarios for exposure of animals and humans to algal or cyanobacterial toxins (specifically the highest quantiles). 
Such information contributes to ecotoxicological studies used to develop thresholds that constitute significant exposure to these toxins~\cite{Mehinto2021}. 
Robotic approaches for quantifying algal biomass across quantiles (often using chlorophyll or chlorophyll fluorescence as a proxy for algal biomass) have become a mainstay for quantitatively documenting heterogeneity in natural ecosystems~\cite{Zhang2020,Sharp2021}.}

Adaptive surveys for selecting collection locations have typically focused on finding high-concentration areas from which to measure, and perform the specimen collection onboard the robot or by a following robot~\cite{das_data-driven_2015, other_onboard_sampling}.
Our work differs from such works as we do not seek to only find maximal concentration areas, \rev{but seek to find a variety of concentration values specified by the quantiles. }

\textbf{Estimation of Quantiles and Quantile Standard Error}
\label{sec:quantile-estimation}has been proposed using statistical techniques.
We estimate the quantile value from measurements using $\estimatedquantilevalue = x_{\lfloor h \rfloor} + (h - \lfloor h \rfloor)(x_{\lceil h \rceil} -x_{\lfloor h \rfloor})$\footnote{\rev{$\lfloor x \rfloor = max\{a \in \mathbb{Z} | a \leq x\}~\textrm{and}~\lceil x \rceil = min\{b \in \mathbb{Z} | b \geq x\}$}}
%\rev{ {$\lfloor x \rfloor = floor(x) $, $\lceil x \rceil = ceiling(x)$}}
where $h = (n - 1)\quantile + 1$, $n$ is the number of measurements, and $\quantile$ is the quantile (Equation 7 in~\cite{hyndman_sample_1996}).
%\footnote{This is the default in the \texttt{numpy} Python package.}.
Standard error of a quantile estimate is a measure of the uncertainty, and is typically used in the construction of confidence intervals of a quantile estimate.
Given a probability density function $p$, the standard error of the $q$th quantile is 
$\sqrt{q(1-q)} / (\sqrt{n}p(\estimatedquantilevalue))$.
Typically $p$ is not known, but can be estimated from samples, using a density estimator~\cite{WILCOX2012103}.
Here we use a Gaussian kernel density estimator when planning.
% Other estimation methods include the \mjci method~\cite{maritz_note_1978}, and the bootstrap or jackknife methods which involve calculating the quantiles over repeatedly sampled subsets of the data. 
% These last two methods can be slow for large datasets and require many iterations to converge.

\textbf{Continuous Non-Convex Gradient-Free Optimization} is used to solve general black-box optimization problems.
\rev{We use these methods to select the analysis locations by minimizing our proposed loss function once the IPP survey is completed \rev{(see \cref{sec:location})}.}
%There have been many proposed approaches to solve such optimization problems. 
% Cross Entropy

The cross-entropy (CE) method is a popular approach which works by maintaining an estimate about the distribution of good solutions, and iteratively updating the distribution parameters based on the quality of the solutions sampled at each step, where the quality is determined via some function of the solution configuration~\cite{de2005tutorial}.
More precisely, a prior $Pr$ and a posterior $Po$ set of samples are maintained.
$Po$ contains $n$ configurations sampled from \rev{$\mathcal{N}(\vec{\mu}$, $\vec{\sigma})$}.
$\vec{\mu}$ and $\vec{\sigma}$ are computed from $Pr$, which is the best $\eta\%$ configurations from $Po$.
This continues iteratively, minimizing the cross-entropy between the maintained and target distributions~\cite{de2005tutorial}.

% Simulated Annealing 
Simulated annealing (SA) \cite{kirkpatrick1987optimization}  is an iterative algorithm inspired by statistical mechanics. SA optimizes an energy function similar to a loss function in other optimization schemes, as configurations with lower energy are preferred. SA begins with a temperature $T = T_\t{max}$. At each iteration, $T$ is decreased exponentially, and the configuration is slightly perturbed randomly. The perturbed state is either accepted or rejected probabilistically based on its energy and $T$. 
%The algorithm terminates when $T$ reaches a given threshold.

%BO
Bayesian optimization (BO) is a method for selecting parameters for difficult optimization problems such as hyperparameter selection~\cite{bohyper}.
BO is similar to IPP and sequential Bayesian optimization~\cite{Marchanta} in that these methods build a model using a GP and select points which maximally improve this model. 
% In pure BO the optimizer is allowed to select any points in a defined boundary, as opposed to IPP and sequential Bayesian optimization, where the robot can only select points it can locally travel to. 
BO uses acquisition functions, such as expected improvement, and iteratively selects the best point to sample from~\cite{jones_efficient_1998, qin_improving_2017}.

% other methods
% Other methods include genetic or evolutionary algorithms \cite{srinivas1994genetic}.
% These algorithms are inspired by biology and generally incorporate some method for crossover between two solutions or mutations to encourage exploration.
%\vspace{-2px}
\section{Formulation}
%\vspace{-1px}
We use a grid-based representation of the planning space, $\gtlocations$, which defines the set of locations that the robot could visit.
For a robot that moves in $\mathbb{R}^d$, $\gtlocations \subset \mathbb{R}^d$.
%$\locations$ is the set of locations that the robot has visited. 
$\gtsensedlocations$ is the set of locations the robot could measure. 
If the robot sensor has finer resolution than $\gtlocations$,
e.g. if the robot uses a camera sensor or takes measurements while traversing between grid points,
then $|\gtsensedlocations| > |\gtlocations|$.
We define \rev{$\sensedlocations_{0:t}$ as the locations the robot has measured up to time $t$} and $\gtsensedvalues$ and \rev{$\sensedvalues_{0:t}$ as the values at all possible measured locations, and the values the robot has measured up to time $t$}, respectively. %, such that $|\gtsensedlocations| = |\gtsensedvalues|$ and $|\sensedlocations| = |\sensedvalues|$.

We define the ground truth quantile values as 
$\quantilevalues = 
quantiles(\gtsensedvalues,\quantiles)$
where $quantiles$ is the function described in \Cref{sec:quantile-estimation} which computes the values $V$ of the quantiles $\quantiles$ of a set of measurements, \rev{in this case $\gtsensedvalues$}.
To define the robot's estimated quantile values, we compute 
$\estimatedquantilevalues = quantiles(\mu(\gtsensedlocations),\quantiles)$
that is, the quantile values of the predicted values from the robot's current GP for all locations the robot could sense.
This is done  to prevent the number of measurements from which the quantile values are estimated from changing as the robot explores (instead of, e.g. using $\mu(\sensedlocations)$).
By doing this, we ensure we always estimate the quantiles across the entire measurable area.
% To assess estimation accuracy, we compute the error using $RMSE (\estimatedquantilevalues , \quantilevalues)$.
During planning, we aim to minimize this error by taking actions which maximize an objective function $\objectivefunction$ that minimizes the error in the quantile value estimate.

To suggest locations for the quantile values, we aim to find a set of $\numtiles$ locations $\quantilespatiallocations$ in the continuous space whose values at those locations are equal to the quantile ground-truth values. 
A set of locations is defined as $\quantilespatiallocations \in \quantilespatiallocations^\#$ where $\quantilespatiallocations^\# \subset \mathbb{R}^{d \times |Q|}$ and $\quantilespatiallocations^\#$ is continuous over the space of $\gtsensedlocations$.
% We define any one set of suggested locations with $\quantilespatiallocations \in \quantilespatiallocations^*$.
% There may be multiple possible such $\quantilespatiallocations$ that satisfy this goal; we are interested in finding one of them.
In practice, the robot only has access to $\estimatedquantilevalues$ (not $\quantilevalues$) during the selection process, so the problem of finding the estimated quantile spatial locations $\estimatedquantilespatiallocations$ can be stated as shown in \cref{eq:point_selection} with some selection loss function $\score$ \rev{(see \cref{eq:ps_loss})}.
\rev{
\begin{equation}\label{eq:point_selection}
    \estimatedquantilespatiallocations = \argmin_{\quantilespatiallocations' \subset \quantilespatiallocations^\#}  \score (\estimatedquantilevalues, \quantilespatiallocations')
\end{equation} 
}

% this vspace is needed only when previous eqn is in \rev{} 
\rev{\vspace{-0.15in}}
\section{Approach}
Figure \ref{fig:hero} illustrates our method.
We separate our approach into two steps: the survey, and the suggestion of locations.
When performing the survey using IPP (\Cref{sec:planning}), the robot takes measurements of the environment to improve its estimate of the quantiles.
After the survey has concluded, location selection (\Cref{sec:location}) \rev{produces} locations for scientists to visit to perform specimen collection.

\subsection{Informative Path Planning}\label{sec:planning}
To plan which locations to measure, the robot uses a POMDP formulation of IPP. 
In order to generate a policy, we use the partially observable Monte Carlo planner (POMCP)~\cite{silver2010pomcp}.
POMCP uses Monte Carlo tree search to create a policy tree. 
To expand the tree and estimate rewards, the tree is traversed until a leaf node is reached.
From the leaf node, \rev{the reward conditioned on that action is estimated by a random policy rollout which is executed until the discounted reward is smaller than some value $\epsilon$.}
We modify the rollout reward to be fixed horizon, giving a reward of zero after a certain number of random policy steps.
We adopt the t-test heuristic for taking multiple steps from a POMCP plan for IPP to improve performance of the planner with fewer rollouts~\cite{denniston_icra_2020}.
We define $GP_{i-1} = GP(\sensedlocations_{0:i-1},\sensedvalues_{0:i-1}; \theta )$
as a GP conditioned on the \rev{previous locations ($\sensedlocations_{0:i-1}$) and measurements ($\sensedvalues_{0:i-1}$)} before measuring a proposed value, and 
$GP_i = GP(\sensedlocations_{0:i-1} \cup \sensedlocations_i, \sensedvalues_{0:i-1} \cup \sensedvalues_i; \theta )$ 
as a GP conditioned on the previous and proposed \rev{locations ($\sensedlocations_i$) and measurements ($\sensedvalues_i$)}, where $\theta$ are GP parameters and $\sensedvalues_i = GP_{i-1}(\sensedlocations_i)$.
Because the observations $GT(x)$ for unseen locations are not known during planning, the predicted mean from \rev{$GP_{i-1}$} is used~\cite{Marchanta}.
%\rev{IPP using GPs suffers from the standard drawbacks of an $O(n^3)$ GP step, which can be alleviated by sparse GP approximations. In our case, \cref{eq:quantile_change,eq:standard_error} each require computing the mean value over $\sensedlocations$ twice.}
%
% In this work, we modify the rollout reward to be a fixed horizon rollout, shown in 
% \begin{equation}\label{eq:max_depth}
%     R(\sensedlocation_i) = 
%      \begin{cases} 
%       0 & i\geq H \\
%       \objectivefunction(\mu(\sensedlocation_i)) 
%   \end{cases}
% \end{equation}
% where $i$ is the depth from the root node, $H$ is the fixed horizon parameter, $\objectivefunction$ is the objective function, and $\mu$ is the predicted output of the GP conditioned on the previous observations at $\location_i$.
% This fixed horizon objective function is used to allow the execution of expensive objective functions.

\textbf{Objective Functions for Quantile Estimation}
We develop two novel objective functions to improve \rev{the quality of quantile value estimates}.
Both compare a measure of the quality of the quantiles estimated by the GP before and after adding a measurement to the GP,
and include an exploration term $c_{plan}\sigma^2(\sensedlocation_i)$ inspired by the upper confidence bound objective function~\cite{Marchanta}, where $c_{plan}$ is a chosen constant.
For both proposed objective functions we use \cref{eq:general_objective}, where $\delta$ is defined by the objective:
\rev{
\begin{equation}\label{eq:general_objective}
    \objectivefunction(\sensedlocations_i) = \frac{\delta(\sensedlocations_i)}{\numtiles} +  \sum_{\sensedlocation_j \in \sensedlocations_i} c_{plan}\sigma^2(\sensedlocation_j), 
\end{equation}}

The first objective function, which we call \textit{quantile change}, is based on the idea of seeking out values which change the estimate of the quantile values by directly comparing the estimated quantiles before and after adding the measured values to the GP.
The idea behind this is that a measurement which changes the estimate of the quantiles indicates that the quantiles are over- or under-estimated.
This can be seen in in \cref{eq:quantile_change}:
% \begin{align}\label{eq:quantile_change}
% \begin{split} 
%   \delta  &= ||quantile(\mu_{PRE}(\gtsensedlocations),\quantiles) - \\
%   & \hphantom{||qua} quantile(\mu_{POST}(\gtsensedlocations),\quantiles) ||_{1} \\
% %   \objectivefunction_\t{qc}(\sensedlocation_i) &= \frac{\delta}{\numtiles} + c \sigma^2(\sensedlocation_i) \\ 
% \end{split}
% \end{align}
\rev{
\begin{align}
\begin{split}\label{eq:quantile_change}
  \delta_\t{qc}(\sensedlocations_i) = \|&quantile(\mu_{GP_{i-1}}(\gtsensedlocations),\quantiles) - \\ &quantile(\mu_{GP_{i}}(\gtsensedlocations),\quantiles) \|_{1} \\
%   \objectivefunction_\t{qc}(\sensedlocation_i) &= \frac{\delta}{\numtiles} + c \sigma^2(\sensedlocation_i) \\ 
\end{split}
\end{align}
}
The second objective function we develop, which we call \textit{quantile standard error}, is based on the change in the estimate of the standard error for the estimated quantiles.
It draws from the same idea that if the uncertainty in the quantile estimate changes after observing a measured value, then it will change the estimate of the quantile values, shown in \cref{eq:standard_error}:
\rev{
\begin{align}\label{eq:standard_error}
\begin{split} 
   \delta_\t{se}(\sensedlocations_i) &= \|se(\mu_{GP_{i-1}}(\gtsensedlocations),\quantiles) - se(\mu_{GP_{i}}(\gtsensedlocations),\quantiles) \|_{1} \\
%   \objectivefunction_\t{se}(\sensedlocation_i) &= \frac{\delta}{\numtiles} + c \sigma^2(\sensedlocation_i) \\
\end{split}
\end{align}
}
$se$ is an estimate of the standard error of the quantile estimate for quantiles $\quantiles$. 
$se$ uses a Gaussian kernel density estimate (we found it faster and more stable than other standard error estimators).
%Other estimation methods include the \mjci method~\cite{maritz_note_1978}, and the bootstrap or jackknife methods which involve calculating the quantiles over repeatedly sampled subsets of the data. 
%These last two methods can be slow for large datasets and require many iterations to converge.
% To compute the objective function over a set of measured points, we average the objective function at each point.

\textbf{Baseline Objective Functions}
We compare against two baselines, one which maximizes spatial coverage of a phenomena and another which seeks maximal areas. 

\textit{Entropy} is a common objective function for IPP when only good spatial coverage of the environment is desired~\cite{kemna_pilot_2018,guestrin_near-optimal_2005,denniston_comparison_2019}. 
It provides a good baseline as it is often used when the specific values of the underlying concentration are unknown.
Entropy is defined \cref{eq:entropy}:
\rev{
\begin{equation}\label{eq:entropy}
   \objectivefunction_\t{en}(\sensedlocations_i) = \sum_{\sensedlocation_j \in \sensedlocations_i}\frac{1}{2} \log(2 \pi e \sigma^2(\sensedlocation_j))
\end{equation}
}

Another objective function we compare against is \textit{expected improvement} (EI), which is widely used in Bayesian optimization and sequential Bayesian optimization for finding maxima~\cite{jones_efficient_1998, qin_improving_2017}.
EI favors actions that offer the best improvement over the current maximal value, with an added exploration term $\xi$ to encourage diverse exploration.
The EI objective function is defined according to \cref{eq:ei}:
\rev{
\begin{equation}\label{eq:ei}
     f_{ei}(\sensedlocations_i) = \sum_{\sensedlocation_j \in \sensedlocations_i} I\Phi(Z) + \sigma(\sensedlocation_j)\phi(Z) \\
\end{equation}
where $Z =\frac{I}{\sigma^2(\sensedlocation_j)}$,  $I = \mu(\sensedlocation_j) - \max(\mu(\gtsensedlocations)) - \xi$, and} $\Phi$ and $\phi$ are the CDF and PDF of the normal distribution, respectively.

\subsection{Location Selection}\label{sec:location}
% describe goal of point selection and how it fits into the overall problem 
Our final goal is to produce a set of $\numtiles$ locations $\quantilespatiallocations$, at which the concentration values will be equal to $\quantilevalues$, the values of the quantiles $\quantiles$. 
The selection process can be done offline as it does not affect planning.
% challenges of it
Finding locations that represent $\quantiles$ is difficult because the objective function over arbitrary phenomena in natural environments will likely be non-convex, and in a real-world deployment, the robot will only have an estimate $\estimatedquantilevalues$ of the $\quantilevalues$ it searches for.

With the location selection problem formulation as in \Cref{eq:point_selection}, \rev{we propose the loss function} 
\begin{equation}\label{eq:ps_loss}
\score(\estimatedquantilevalues, \estimatedquantilespatiallocations) = \|\estimatedquantilevalues - \mu(\estimatedquantilespatiallocations)\|_{2} + c_{select} \sigma^2(\estimatedquantilespatiallocations),
\end{equation}
where $c_{select} \sigma^2(\estimatedquantilespatiallocations)$ is an added penalty for choosing points that the GP of collected measurements is not confident about\footnote{\rev{The parameters $c_{plan}$ and $c_{select}$ are distinct.}}, \rev{and $\score$ can be used in any optimization scheme}.
%; $c_{plan}$ is an exploration bonus for reducing uncertainty during the planning phase, while $c_{select}$ is a penalty for choosing uncertain points during the location selection phase.}}.
%
\rev{During optimization, \cref{eq:ps_loss} is evaluated using} $\estimatedquantilevalues$ and returns the suggested specimen collection locations $\estimatedquantilespatiallocations$. 
We compare three \rev{optimization} methods in our experiments to determine which best minimizes \rev{our selection loss function} (\cref{eq:ps_loss}).
A strength of these types of optimization methods is that the formulation allows for suggesting points that may be spatially far from locations the robot was able to measure if they have values closer to the quantile values of interest.

\begin{figure}[t]
    \centering
        \begin{subfigure}{\threeboxplot}
                \centering
                \includegraphics[width=\textwidth,height=\boxplotheight,trim={.8cm 0 .5cm 0},clip]{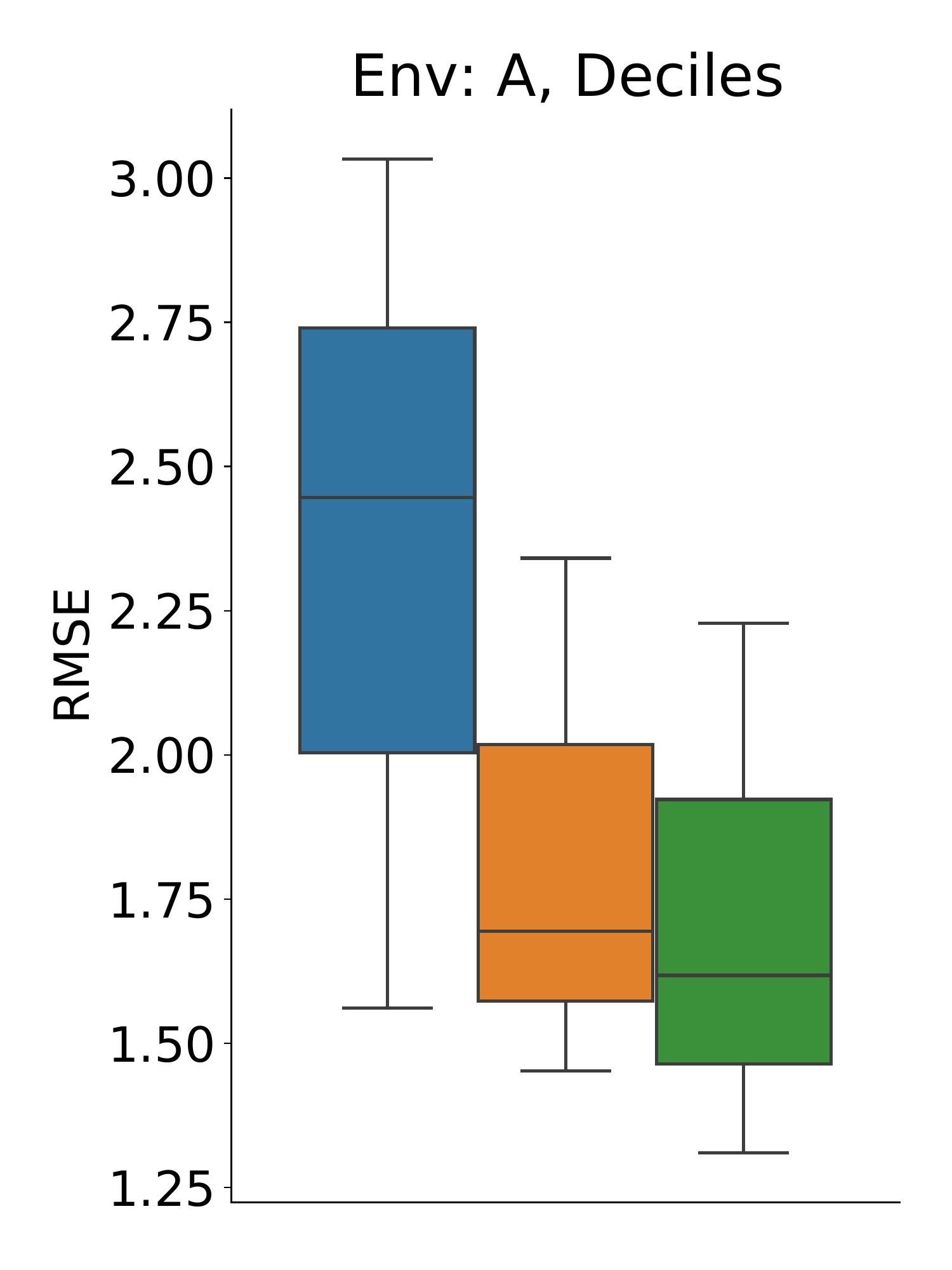}
        \end{subfigure}
        \begin{subfigure}{\threeboxplot}
                \centering
                \includegraphics[width=\textwidth,height=\boxplotheight,trim={.8cm 0 .5cm 0},clip]{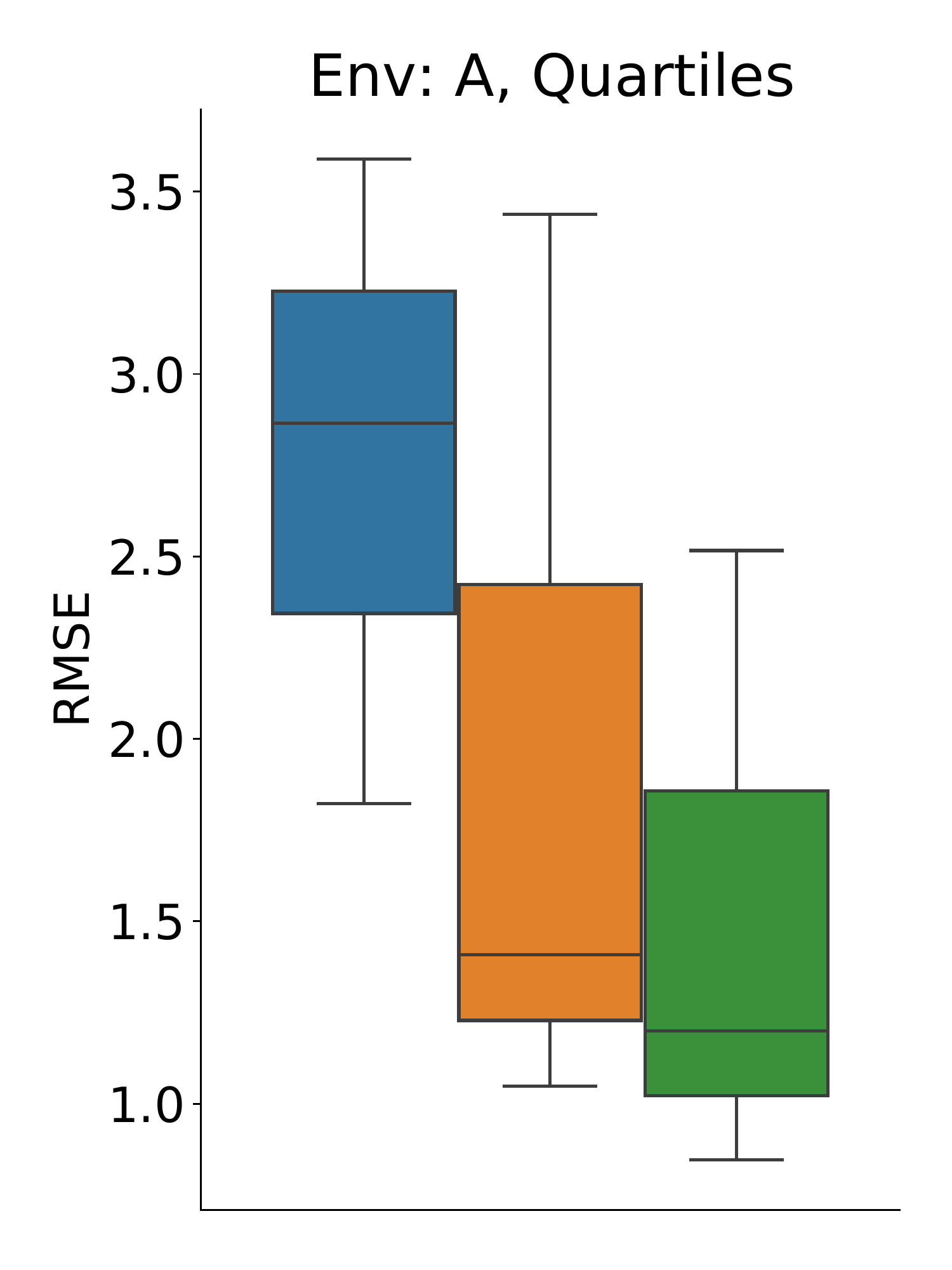}
        \end{subfigure}
        \begin{subfigure}{\fourboxplot}
                \centering
                \includegraphics[width=\textwidth,height=\boxplotheight,trim={.8cm 0 .5cm 0},clip]{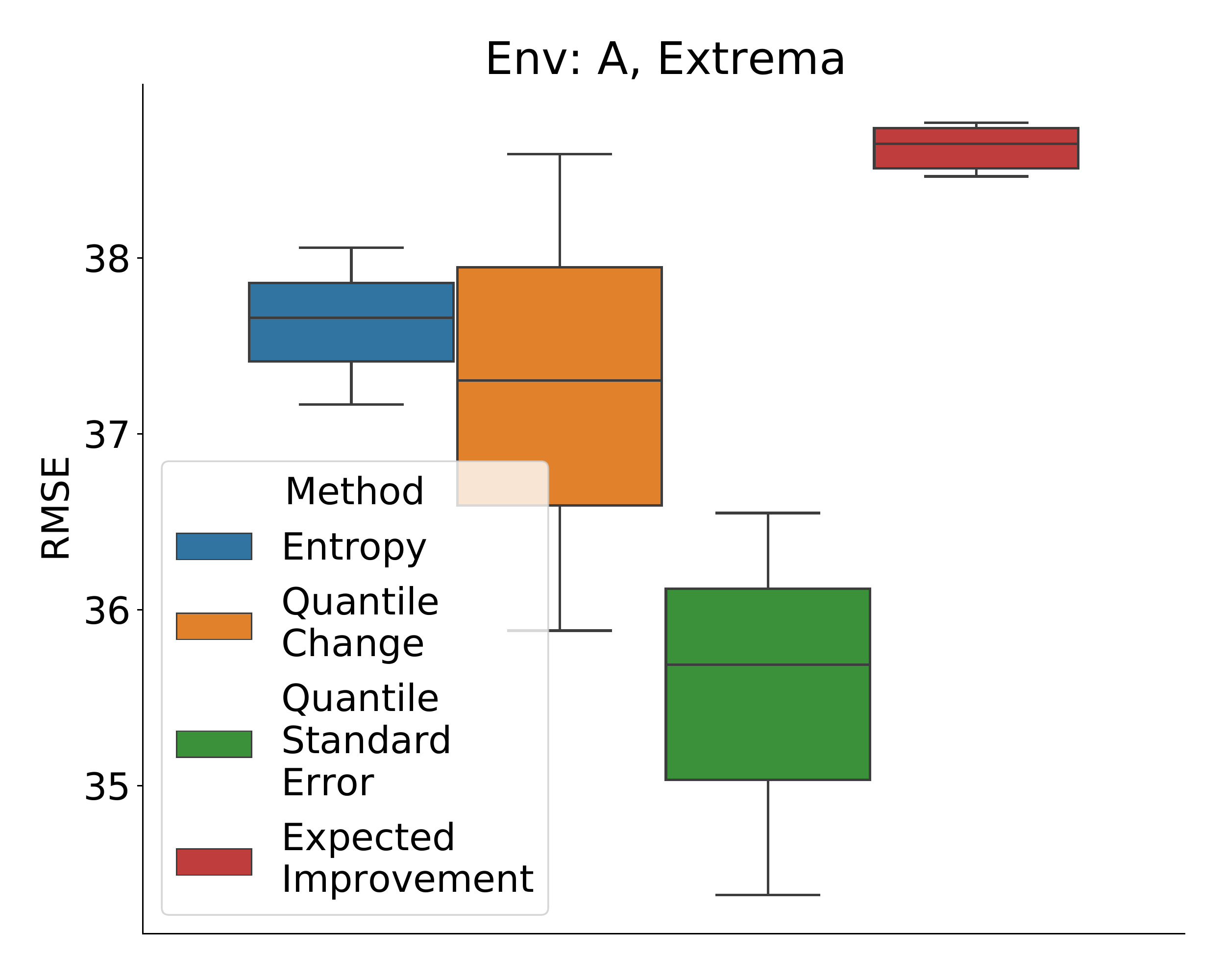}
        \end{subfigure}
        \begin{subfigure}{\threeboxplot}
                \centering
                \includegraphics[width=\textwidth,height=\boxplotheight,trim={.8cm 0 .5cm 0},clip]{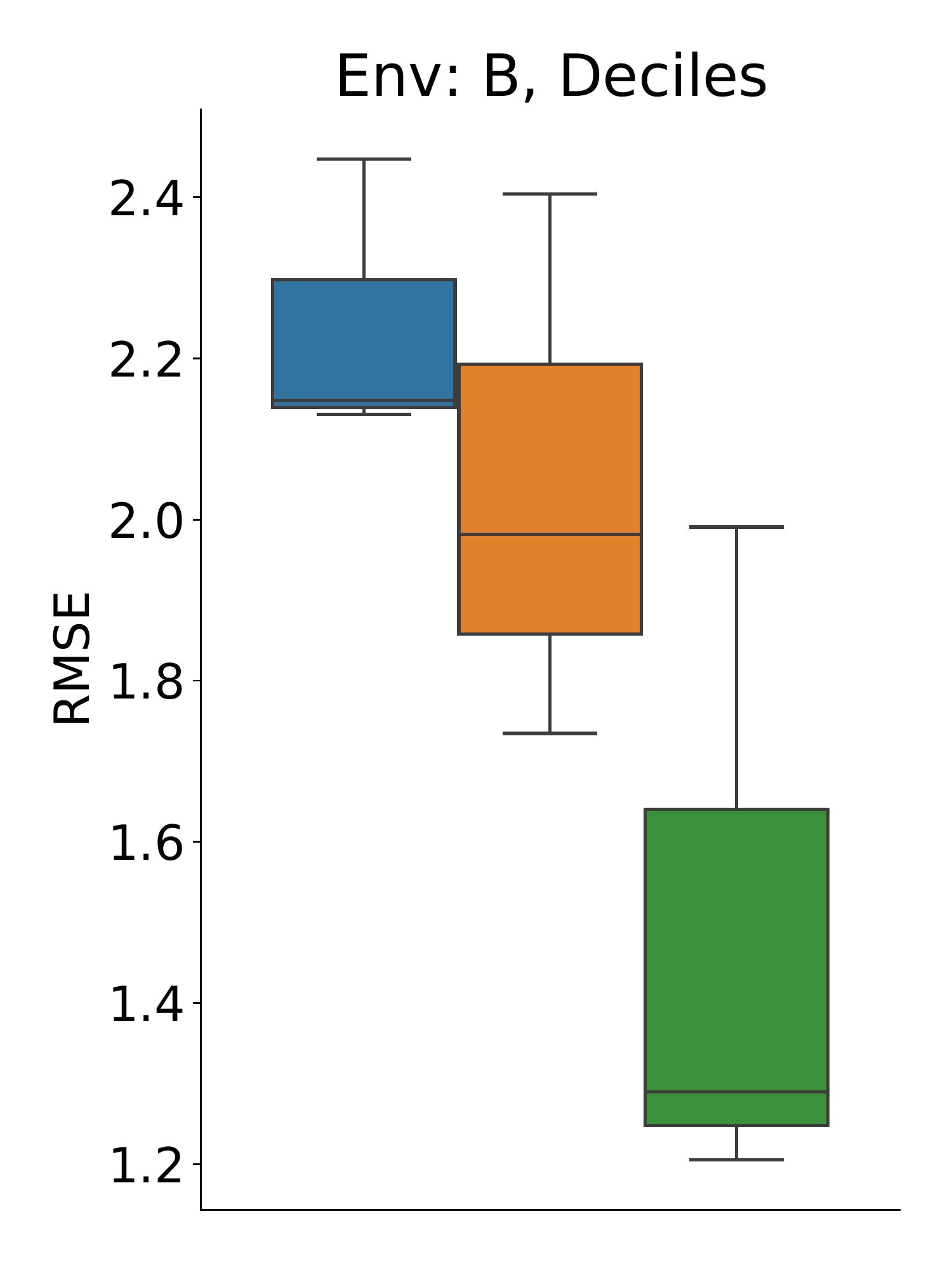}
                
        \end{subfigure}
        \begin{subfigure}{\threeboxplot}
                \centering
                \includegraphics[width=\textwidth,height=\boxplotheight,trim={.8cm 0 .5cm 0},clip]{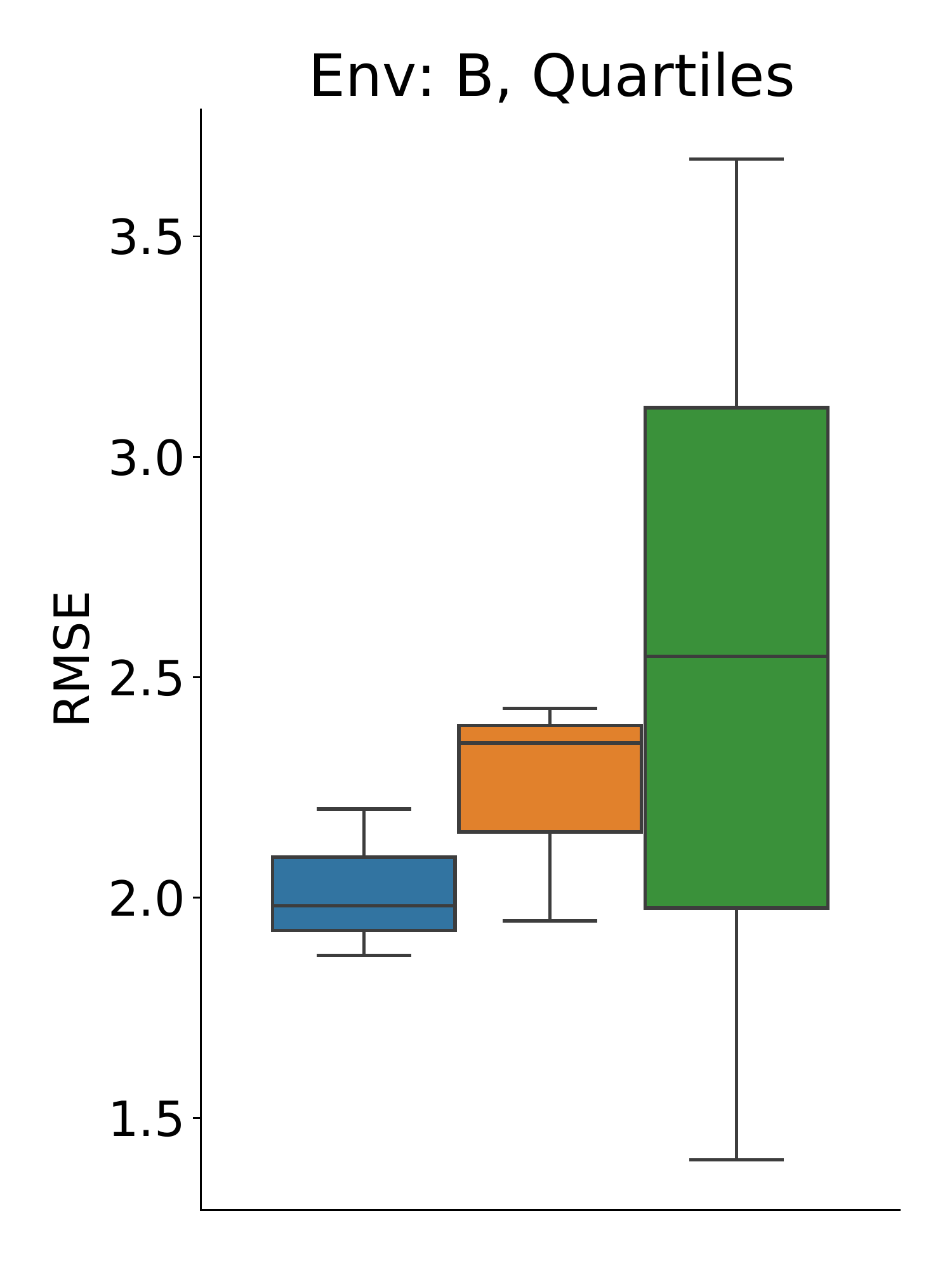}
        \end{subfigure}
        \begin{subfigure}{\fourboxplot}
                \centering \includegraphics[width=\textwidth,height=\boxplotheight,trim={.8cm 0 .5cm 0},clip]{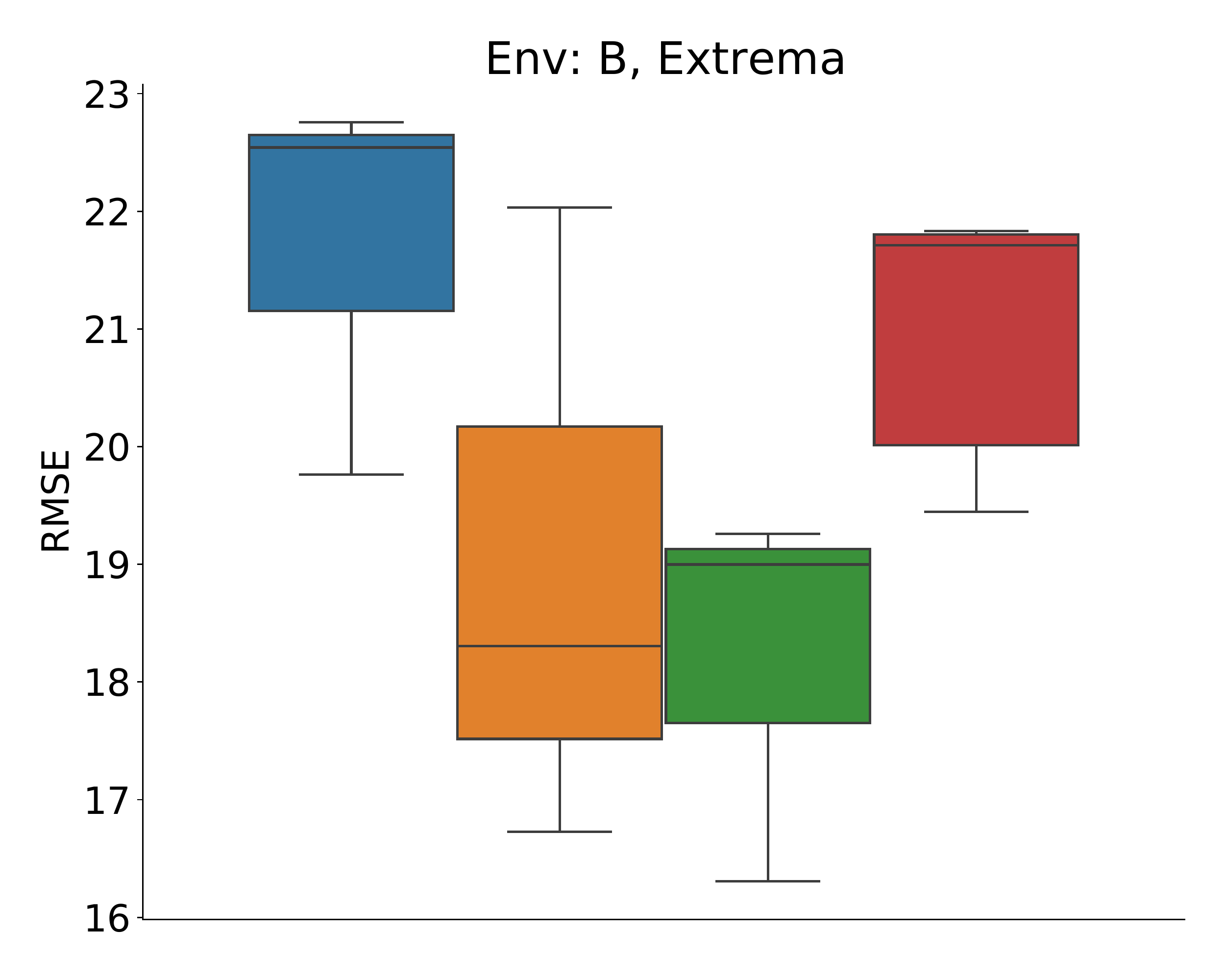}
        \end{subfigure}
        \caption{\textbf{Simulated drone planning experiments with real data.} Error between ground truth quantile values and estimated quantile values. Datasets A and B collected using a hyperspectral camera in Clearlake, California.
        Units are \rev{400nm channel pixel intensity ($0-255$).  \rev{Each dataset and objective function pairing is run three times.}}
        }\label{fig:clearlake_planning_results}
        \vspace{-0.2in}
\end{figure}

\begin{figure}[b!]
    \centering
    \includegraphics[trim={0.5cm .2cm 0.5cm 0.9cm}, clip, width=.45\columnwidth,height=2.6cm]{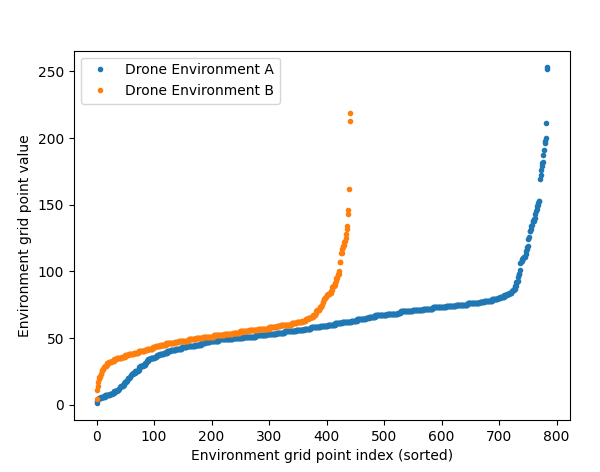}
    \includegraphics[trim={0.5cm .2cm 0.7cm 0.9cm}, clip, width=.45\columnwidth,height=2.6cm]{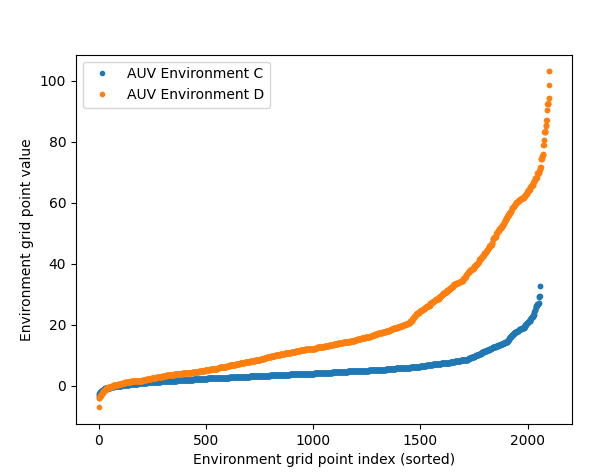}
    \caption{
    \rev{Experimental dataset distributions. Drone data is measured in pixel intensity; AUV data in $\mu g/L$ chlorophyll.}
    }
    \label{fig:environment_distributions}
    \vspace{-7pt}
\end{figure}

% Simulated annealing is a local search method which seeks to improve a single solution by moving to local solutions.
% The starting solution is selected heuristically. 
% This heuristic is the set of points the robot measured that are closest in value to $\estimatedquantilevalues$ for each quantile.

% Bayesian optimization and cross-entropy are both global search methods which maintain a distribution over the search space and update the distribution iteratively.
% These methods differ in the type of distribution over solutions they maintain and their search strategy.

\section{Experiments}
To evaluate our \rev{approaches for planning and for location selection}, we compare against baselines in two different informative path planning tasks in simulation using \rev{four} datasets collected in the real world.

In the first task, a \rev{simulated drone with a virtual camera gathers data from} orthomosaics (a single image produced by combining many smaller images, called orthophotos) collected of a lake using a hyperspectral sensor. 
The orthomosaics, A and B, are taken in the same location but on different days and times.
The drone collects many measurements from one location, \rev{where each is} a pixel in a downsampled image. 
As a proxy for chlorophyll concentration, we measure the \rev{400nm channel} pixel intensity \rev{($0-255$)}.
The drone maintains a constant altitude and moves in a 2D plane with a north-fixed yaw, moving in either the $x$ or $y$ direction per step.

In the second task, an autonomous underwater vehicle (AUV) explores an environment.
\rev{Two AUV surveys, C and D, which} were taken in the same reservoir \rev{but }at different times and different areas, were conducted \rev{in a 3D lawnmower pattern} using a chlorophyll sensor
and are interpolated using a GP to $\gtsensedlocations$.
At each step, the AUV moves in one $x$, $y$, or $z$ direction and takes five evenly spaced measurements when moving \rev{between locations}.

\rev{We evaluate each task on their two respective datasets (A/B, C/D)} and three different quantiles: deciles $(0.1, 0.2, \dots , 0.8, 0.9)$, quartiles $(0.25,0.5,0.75)$, and upper extrema $(0.9,0.95,0.99)$. \rev{See \Cref{fig:environment_distributions} for a summary of the dataset distributions.}

\begin{figure}[t]
    \centering
                \begin{subfigure}{\threeboxplot}
                \centering
                \includegraphics[width=\textwidth,height=\boxplotheight,trim={.8cm 0 .5cm 0},clip]{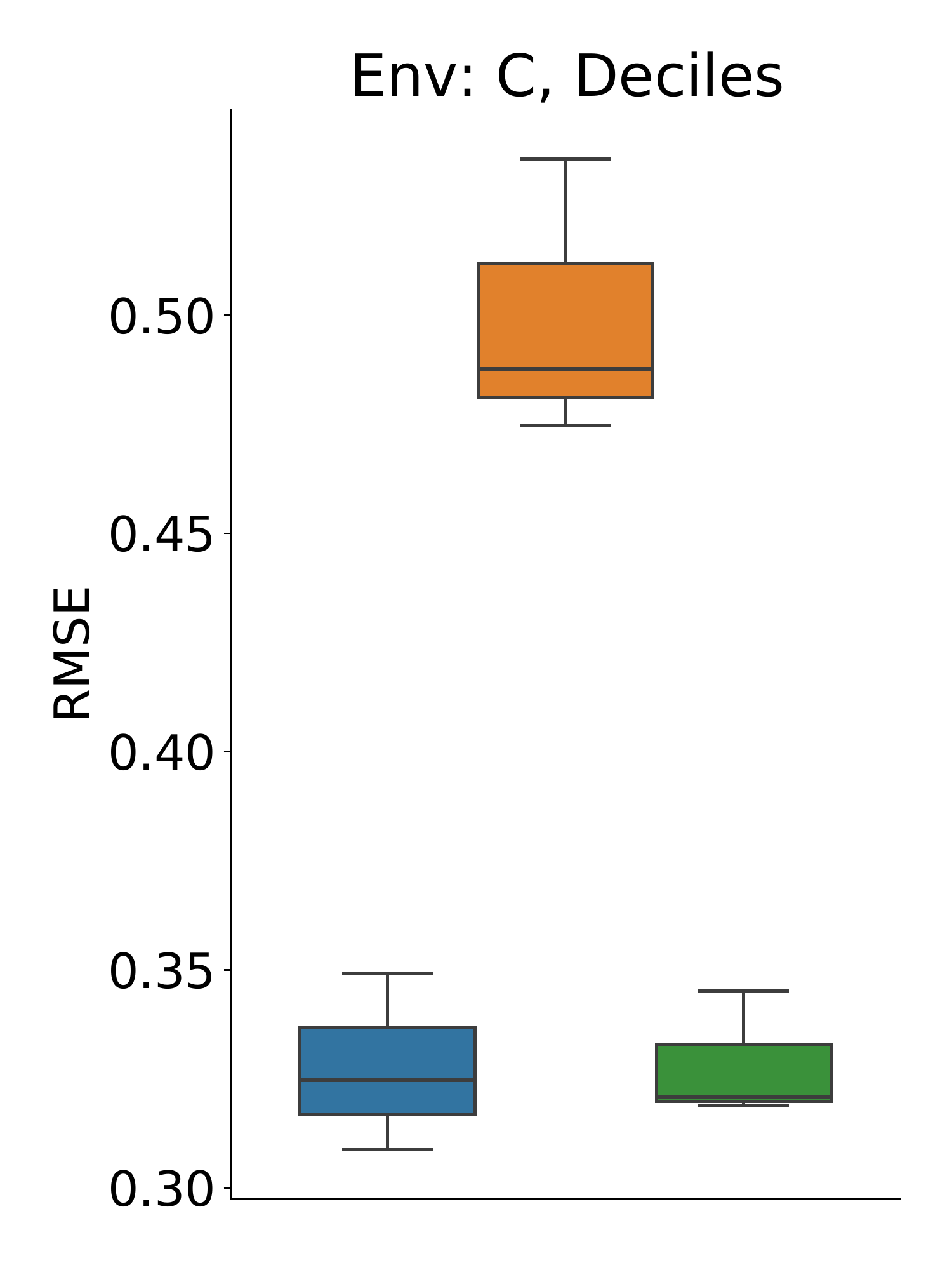}
        \end{subfigure}
        \begin{subfigure}{\threeboxplot}
                \centering
                \includegraphics[width=\textwidth,height=\boxplotheight,trim={.8cm 0 .5cm 0},clip]{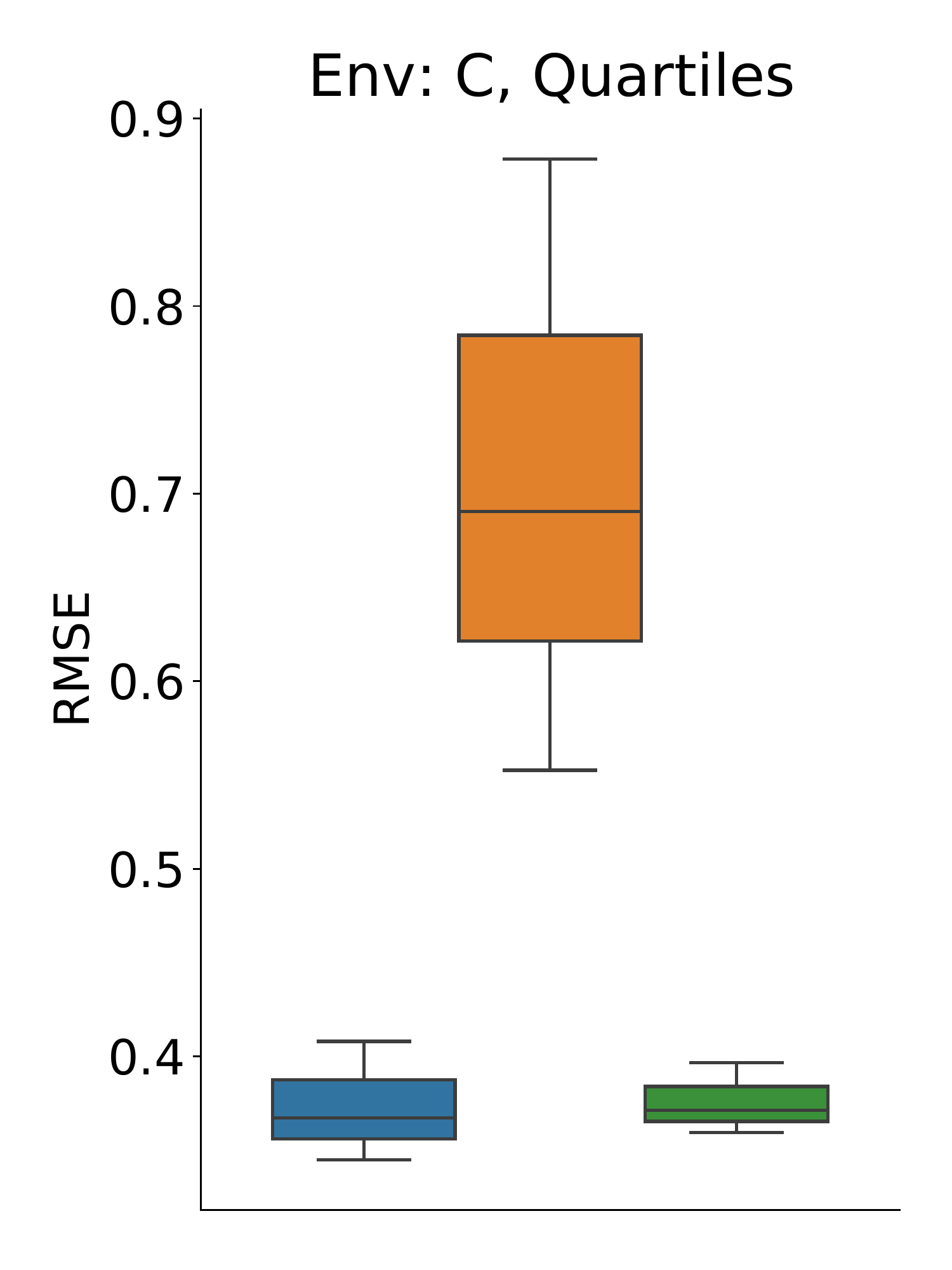}
        \end{subfigure}
        \begin{subfigure}{\fourboxplot}
                \centering
                \includegraphics[width=\textwidth,height=\boxplotheight,trim={.8cm 0 .5cm 0},clip]{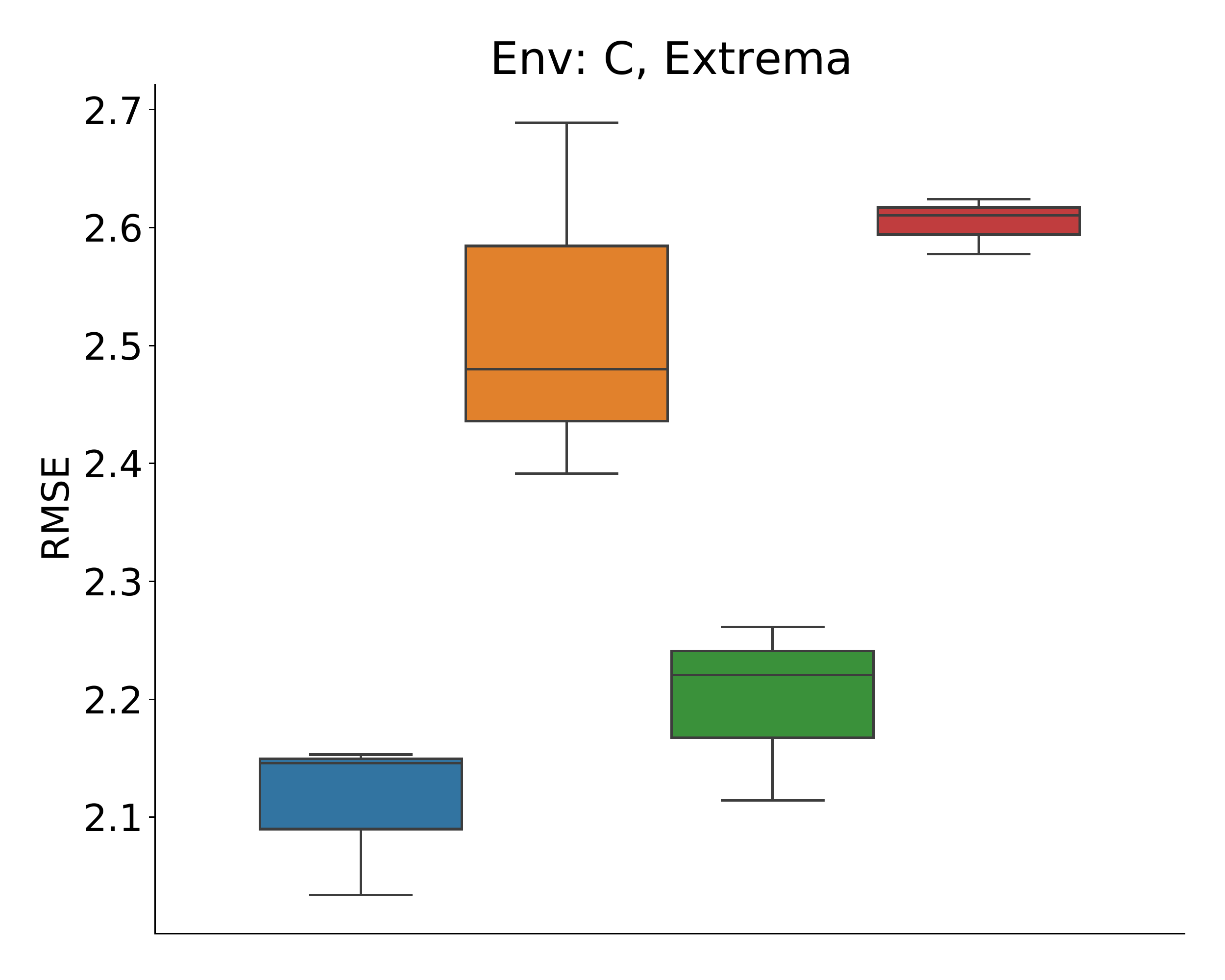}
        \end{subfigure}
        \begin{subfigure}{\threeboxplot}
                \centering
                \includegraphics[width=\textwidth,height=\boxplotheight,trim={.8cm 0 .5cm 0},clip]{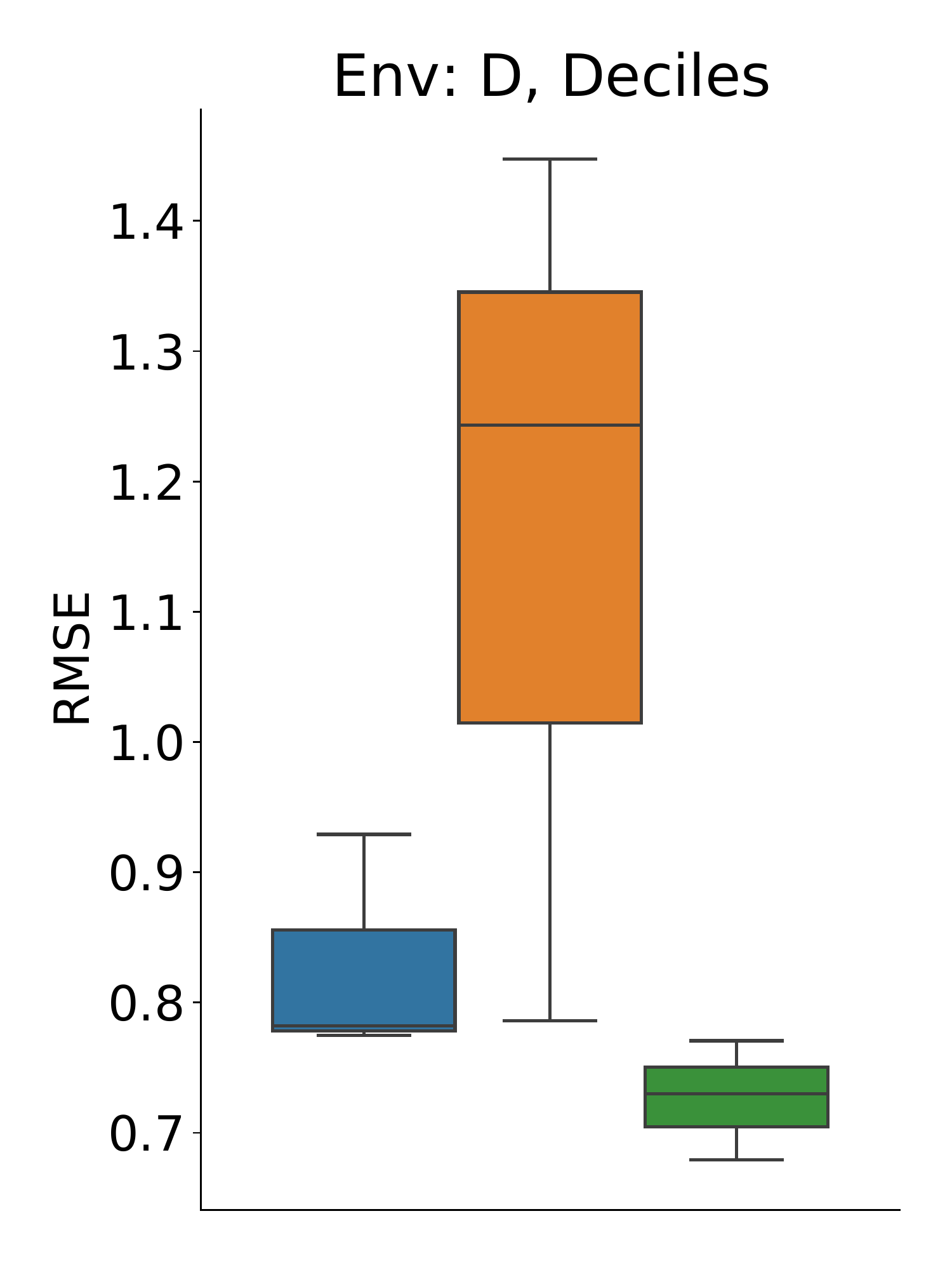}
                
        \end{subfigure}
        \begin{subfigure}{\threeboxplot}
                \centering
                \includegraphics[width=\textwidth,height=\boxplotheight,trim={.8cm 0 .5cm 0},clip]{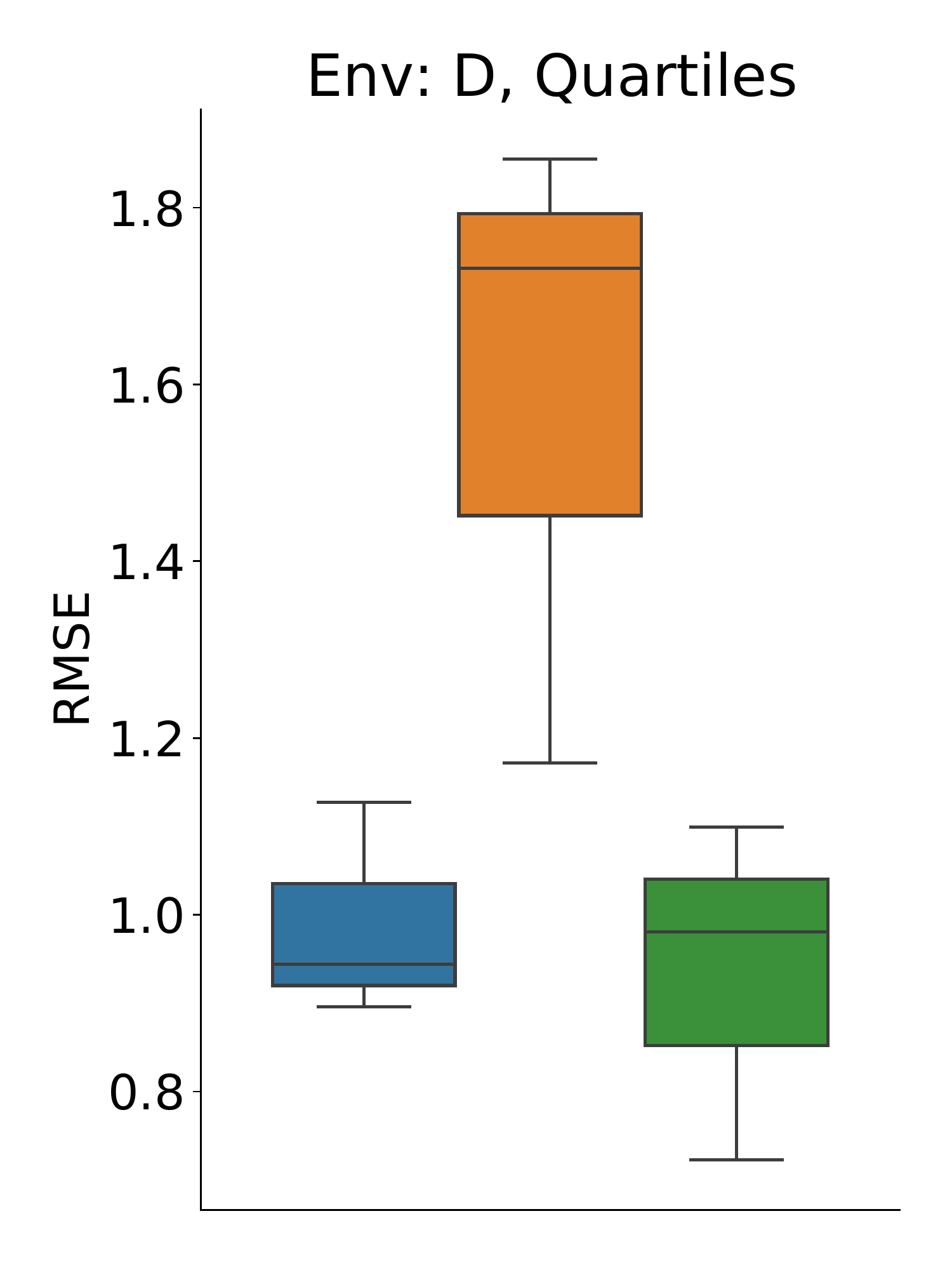}
        \end{subfigure}
        \begin{subfigure}{\fourboxplot}
                \centering \includegraphics[width=\textwidth,height=\boxplotheight,trim={.8cm 0 .5cm 0},clip]{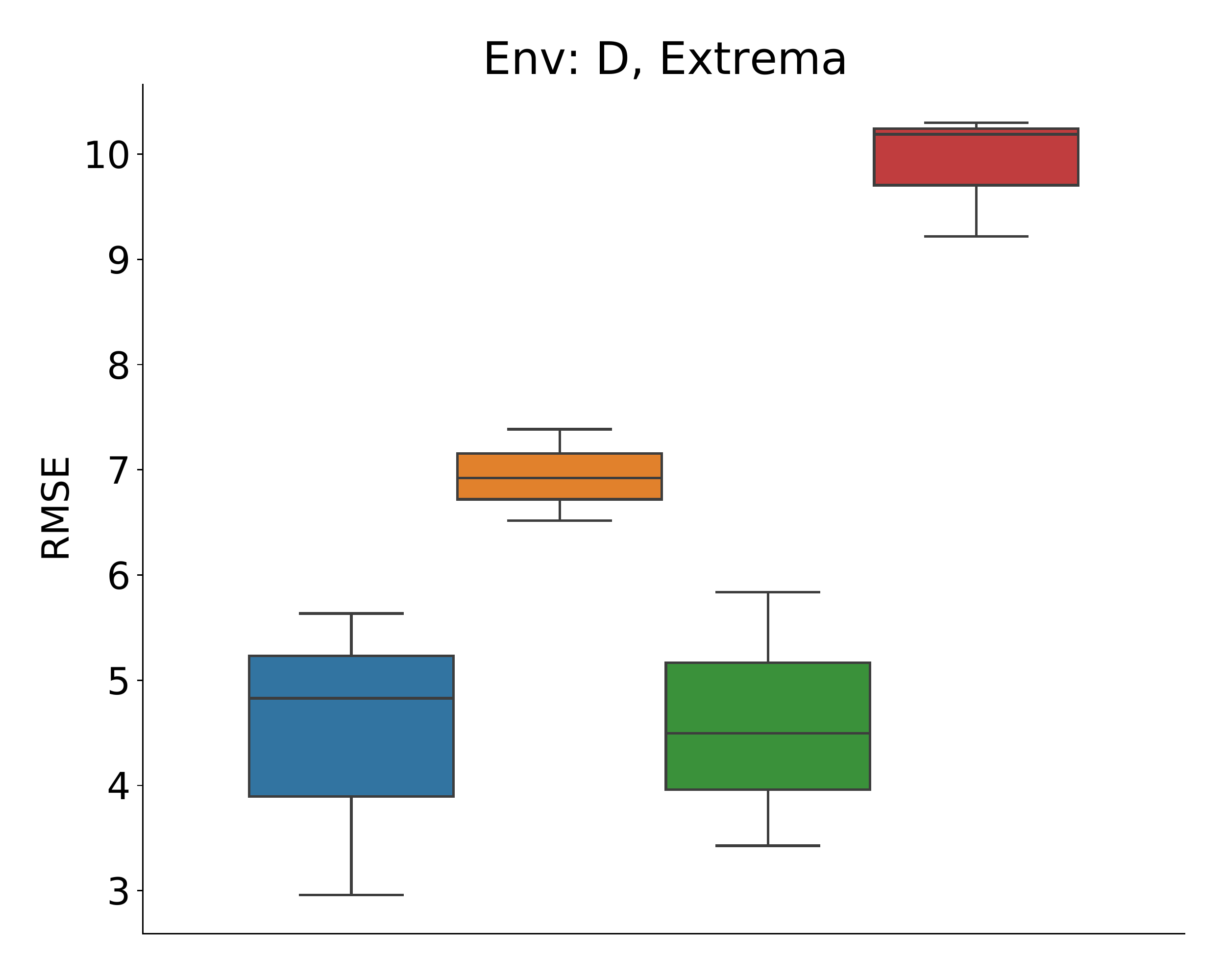}
        \end{subfigure}
        \caption{\textbf{Simulated AUV planning experiments with real data.} Error between ground truth quantile values and estimated quantile values. Datasets C and D collected from a reservoir in California using an underwater robot with a chlorophyll fluorescence sensor.
        Units are $\mu g/L$ chlorophyll.
        \rev{Each dataset and objective function pairing is run three times.}}\label{fig:ecomapper_planning_results}
         \vspace{-0.2in}
\end{figure}

\begin{figure}[b]
    \centering
    \includegraphics[width=0.8\columnwidth,height=2.4cm,trim={.2cm .5cm .5cm .5cm},clip]{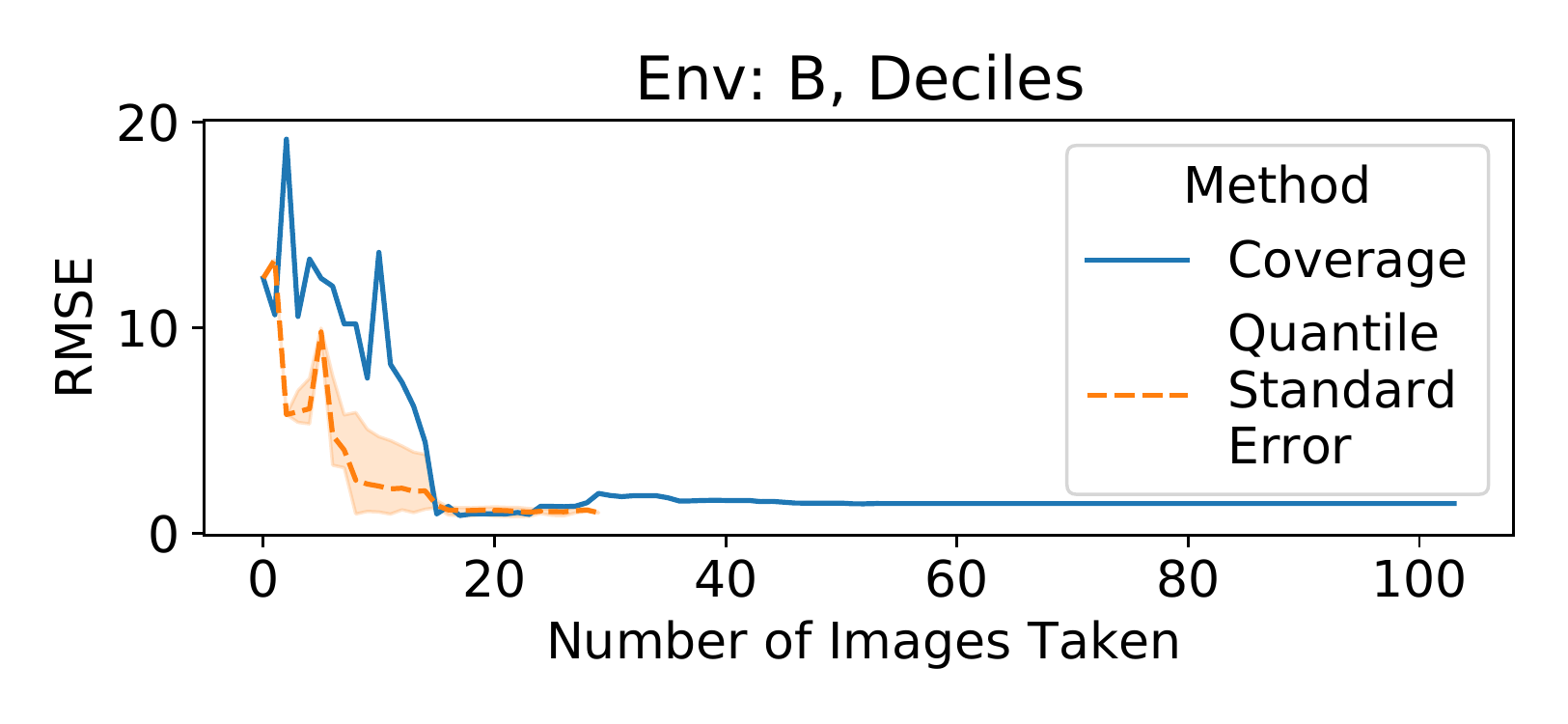}
    \caption{\rev{Comparison between quantile standard error with POMDP solver and Spiral-STC coverage planner~\cite{coverage} on Drone Environment B estimating deciles, each over 3 trials.}}
    \label{fig:coverage_expr}
    \vspace{-7pt}
\end{figure}

\subsection{Informative Path Planning: Objective Functions}
\label{ssec:ipp_experiments}
To evaluate \rev{how well our proposed IPP objective functions estimate the ground truth quantile values in real environments}, we compare quantile change (\cref{eq:quantile_change}) and quantile standard error (\cref{eq:standard_error}) against a baseline entropy objective function (\cref{eq:entropy}).
For the upper extrema quantiles, we also compare against expected improvement (\cref{eq:ei}), as it is similar to a sequential Bayesian optimization based IPP task.

\subsubsection{\rev{Setup}}
In the planner, we use $\gamma = 0.9$, and each trial is run over 3 seeds.
The objective $c_{plan}$ parameter is set to the approximate magnitude of the rewards seen for each environment, which we found experimentally to be an adequate value.
\rev{The GP mean was set to 0 and the datasets were normalized, while the lengthscale was set to 12. The exploration constant $c_{plan}$ was set to $1E-6$ for quantile change and $1E-2$ for quantile standard error.}
\rev{To compare the performance of the approaches, we use the RMSE between the ground truth quantiles, $\quantilevalues$, and estimated quantile values after performing a survey, $\estimatedquantilevalues$. 
% We choose this metric rather than, e.g., comparing the error between the estimated concentration $\mu(\gtsensedlocations)$ and the ground truth concentration  $GT(\gtsensedlocations)$ because the objective of the task specifically focuses on getting the highest accuracy for the quantile values, and does not optimize for overall concentration accuracy.
}

\textbf{Drone with Camera}
The drone is allowed to take 30 simulated pictures out of a grid with about 300 positions. 
Each picture is downsampled to $8\times 5$ pixels ($40$ measurements) with $37.1^{\circ}$ by $27.6^{\circ}$ field of view, similar to the drone used in the field trial reported in \cref{sec:field_trial}.
%At each step, the drone collects $40$ measurements.
\rev{While this is downsampled from the true image, it captures the coarse trends of the concentration, which is both still scientifically useful and performant for GP evaluations.}
For each trial, the GP is seeded with 100 evenly spaced \rev{measurements} across the workspace, as \rev{prior} data. %given by a remote sensor, such as a satellite.
\rev{The planner uses} $300$ rollouts per step, and the maximum planning depth is 7.
%For quantile change $c=10^{-6}$ and for quantile standard error $c=10^{-3}$.

\textbf{AUV with Chlorophyll Sensor}
The AUV is simulated for 200 steps in a $12 \times 14 \times 2$ grid.
The planner uses $130$ rollouts per step and a maximum depth of $10$.
The GP is seeded with \rev{measurements} from 50 locations.

\subsubsection{\rev{Discussion}}
Overall, our results are robust across multiple experimental environments as well as robot sensor types.
We find that planning with the quantile standard error objective function has a 10.2\% mean reduction in median error across all environments when compared to using entropy. 
\rev{This shows that an objective function tailored to estimating a set of quantiles will outperform a coverage planner, such as entropy, that would typically achieve low overall error in reconstructing the entire environment model.}

\begin{figure}[t]
    \centering
        \begin{subfigure}{\thirdcolumn}
                \centering
                \includegraphics[width=\textwidth,height=\boxplotheight,trim={.8cm 0 .5cm 0},clip]{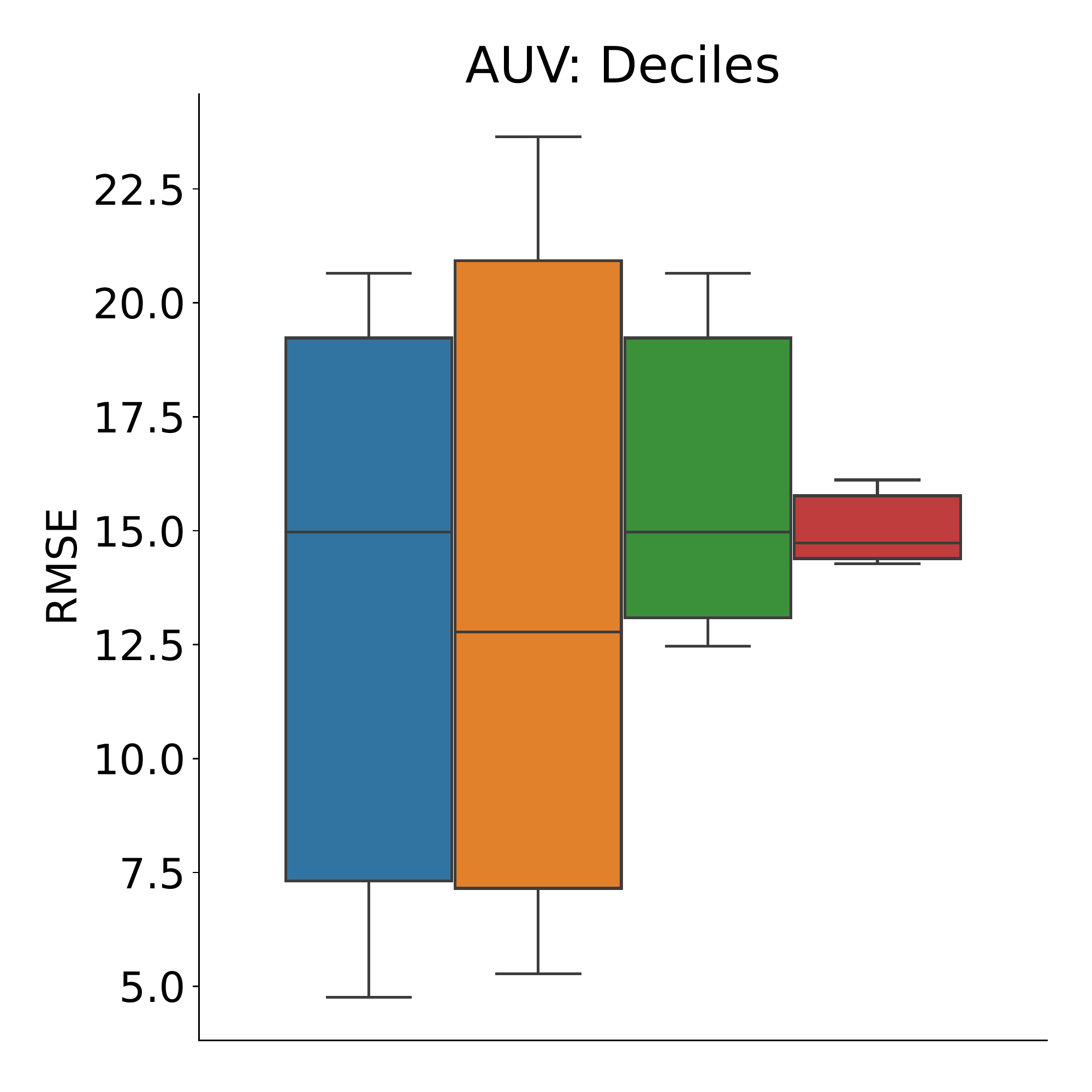}
        \end{subfigure}
        \begin{subfigure}{\thirdcolumn}
                \centering
                \includegraphics[width=\textwidth,height=\boxplotheight,trim={.8cm 0 .5cm 0},clip]{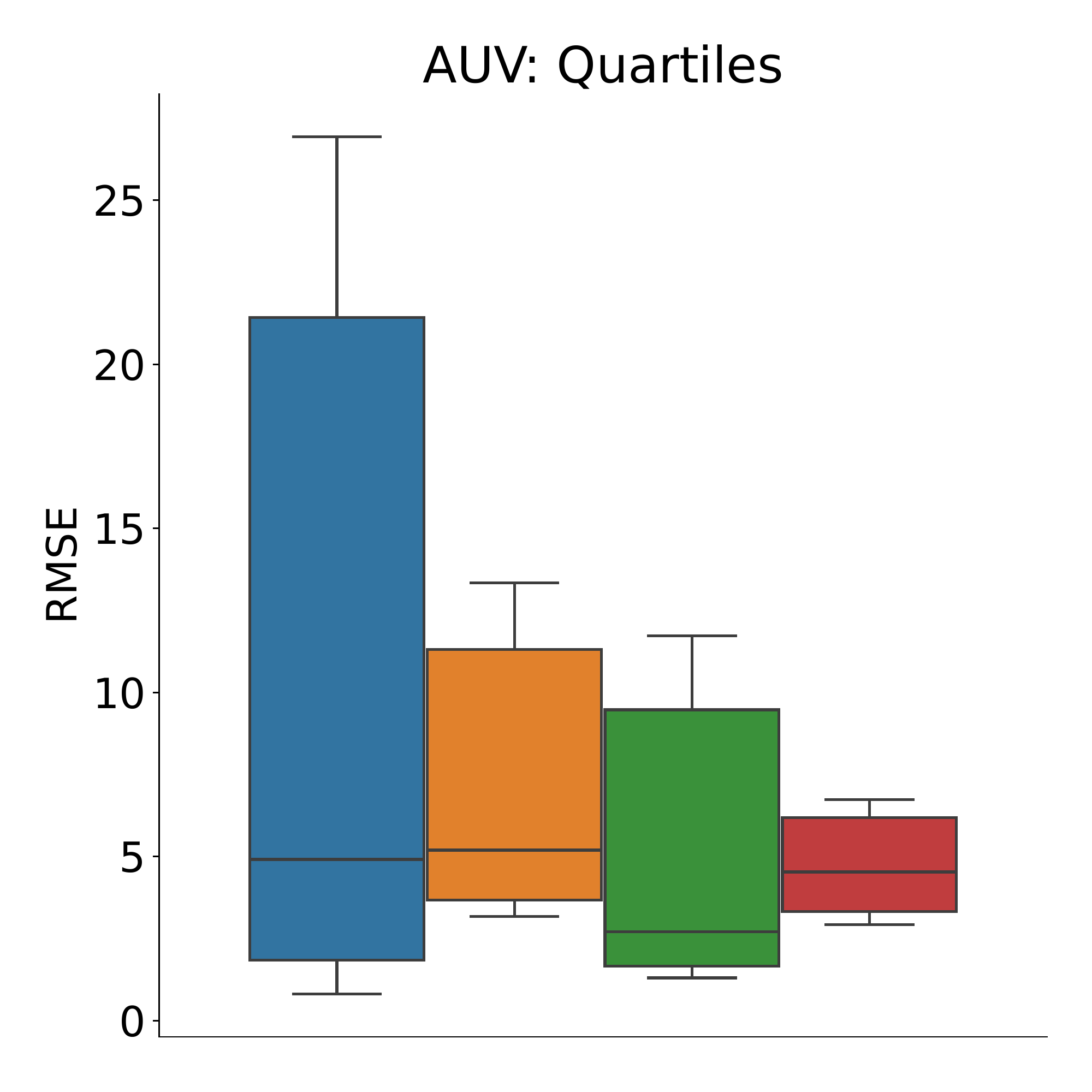}
        \end{subfigure}
        \begin{subfigure}{\thirdcolumn}
                \centering
                \includegraphics[width=\textwidth,height=\boxplotheight,trim={.8cm 0 .5cm 0},clip]{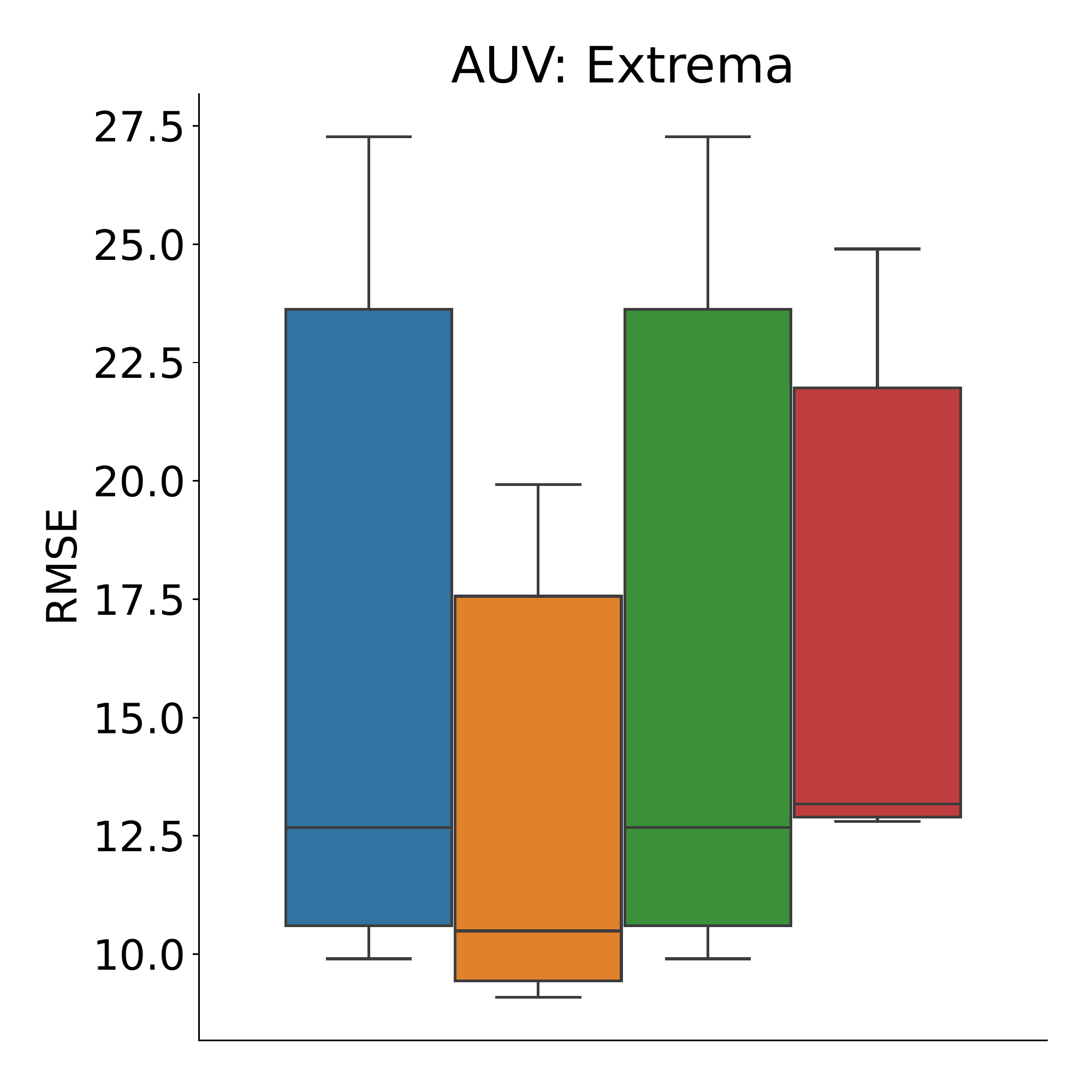}
        \end{subfigure}
                \begin{subfigure}{\thirdcolumn}
                \centering
                \includegraphics[width=\textwidth,height=\boxplotheight]{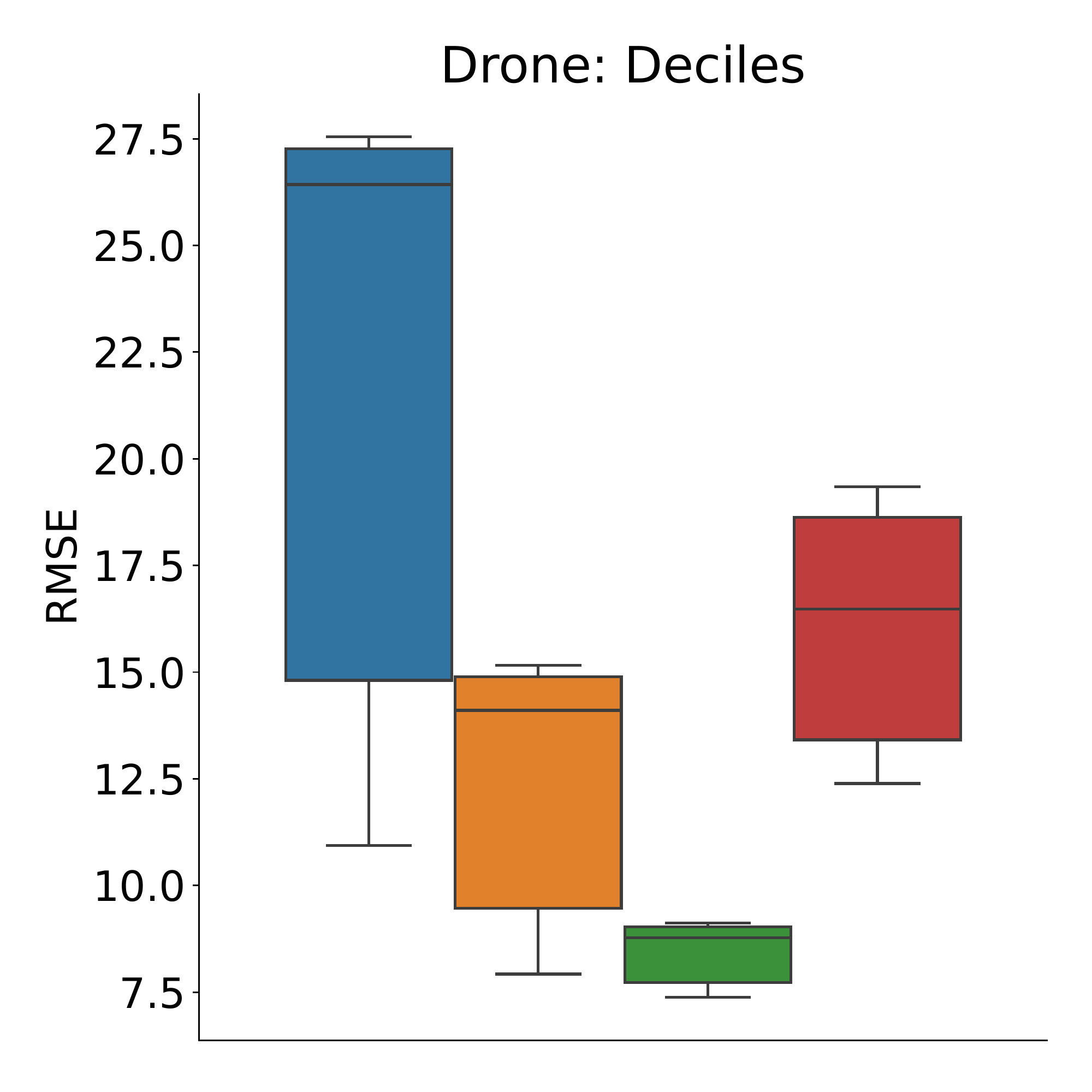}
        \end{subfigure}
        \begin{subfigure}{\thirdcolumn}
                \centering
                \includegraphics[width=\textwidth,height=\boxplotheight]{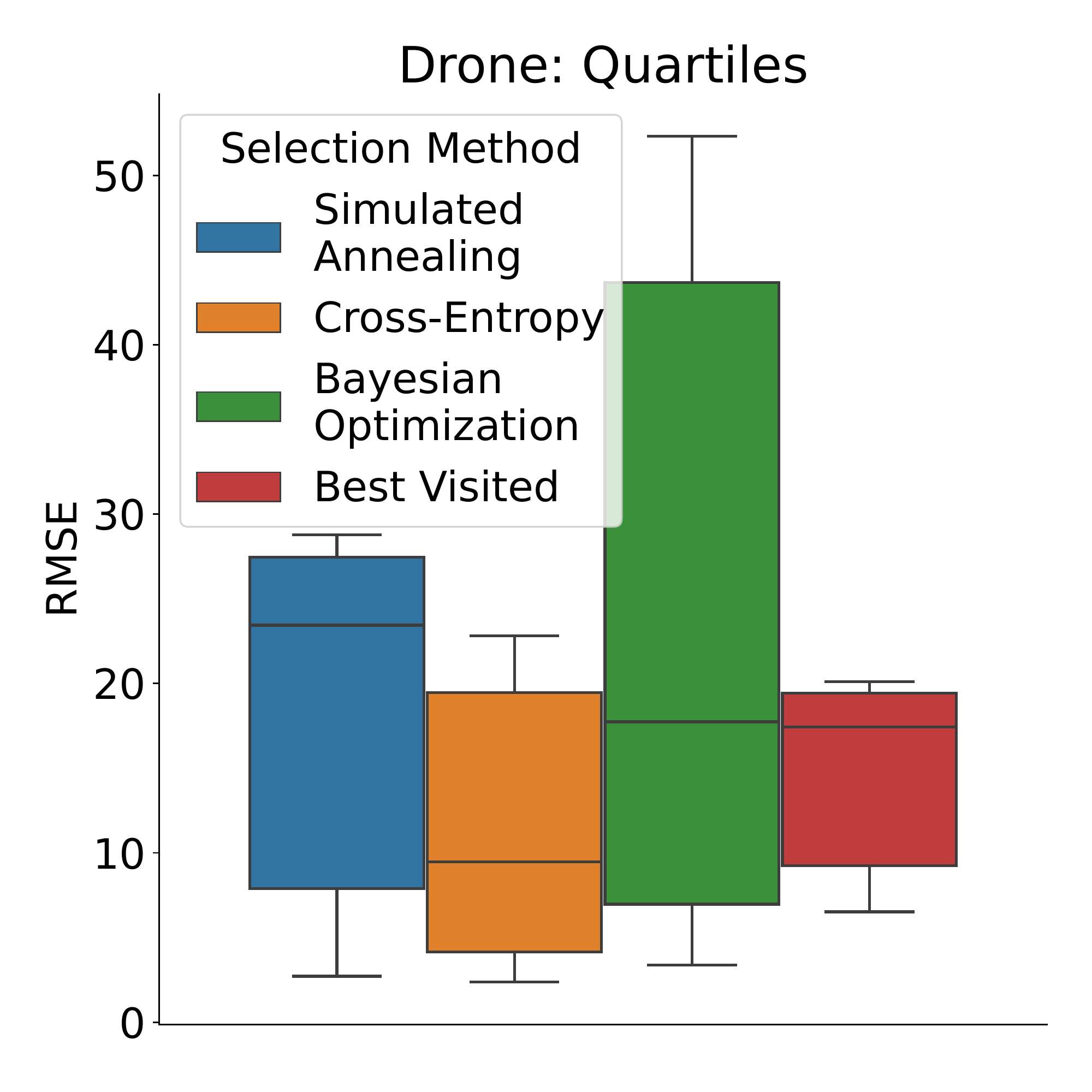}
        \end{subfigure}
        \begin{subfigure}{\thirdcolumn}
                \centering
                \includegraphics[width=\textwidth,height=\boxplotheight]{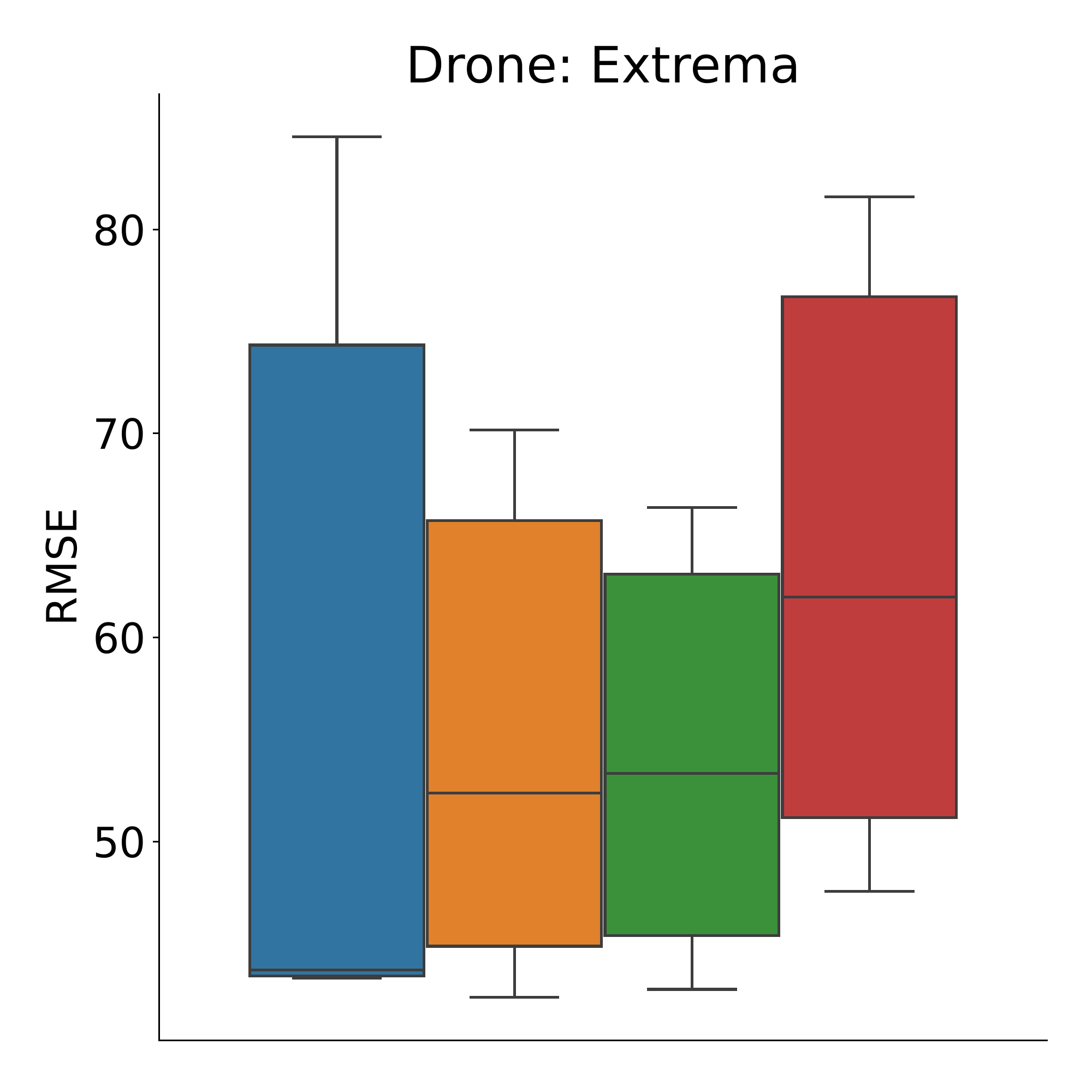}
        \end{subfigure}
        
        \caption{\textbf{Quantile location selection results.} 
        RMSE between the ground truth values at the selected locations and the corresponding true quantile values.
        Units are $\mu g/L$ chlorophyll for AUV, \rev{400nm channel pixel intensity ($0-255$) for Drone}.
     }
     \label{fig:ecomapper_ps_boxplots}
    \vspace{-0.2in}
\end{figure}

\Cref{fig:clearlake_planning_results,fig:ecomapper_planning_results} show the results of planning with the proposed objective functions for the drone and AUV experiments, respectively. 
In the drone \rev{tasks} (A/B),
quantile change and quantile standard error outperform the baseline entropy in estimating the deciles and upper extrema.
In quartiles in environment A, both proposed methods perform well, but in environment B entropy outperforms our methods. 
% We believe this because this specific configuration is the most difficult task as each method performs worse at the end than they do at 5 steps.
For the extrema, both methods perform better than expected improvement, and entropy outperforms expected improvement in environment A. 
We believe this is because expected improvement \rev{focuses explicitly on improving the (single)} maximal value and does not do a good job of localizing high concentration areas, thus \rev{overestimating} the quantile values.

For the AUV \rev{tasks} (C/D),
quantile standard error outperforms entropy in environment C when estimating deciles and quartiles, and performs equally well as entropy in all other tasks besides estimating the extrema in environment D.
Expected improvement performs poorly in estimating the extrema due to similar issues as with the drone.
Quantile change performs poorly in most AUV tasks \rev{in contrast to the drone tasks, where it performs comparably. 
This may indicate that quantile change performance decreases with point sensors, or it is more sensitive to variation in the absolute measurement values.
Thus, we recommend using quantile standard error unless there are computational constraints, since its performance is more consistent.}

\rev{We also compare our approach using the proposed quantile standard error objective function to the Spiral-STC coverage planner~\cite{coverage}. 
\Cref{fig:coverage_expr} shows that the proposed adaptive solution is able to reduce the error between the estimated quantiles and the ground truth quantiles more rapidly than the non-adaptive coverage planner.}

\subsection{Location Selection}
\rev{To evaluate how close the values at locations suggested by our proposed loss function are to the estimated quantile values,} we use the optimization algorithms simulated annealing, cross-entropy, and Bayesian optimization 
with the results from the quantile estimation task \rev{(\Cref{ssec:ipp_experiments}) using the quantile standard error objective function}.
% CE gamma is eta in this paper to avoid confusion with the planner gamma
\subsubsection{\rev{Setup}}
We compute the error by $RMSE(\quantilevalues,\estimatedquantilespatiallocations) = \|(\quantilevalues - GT(\estimatedquantilespatiallocations))\|_2$.
We set $c_{select}$ to 15 for deciles, 200 for quartiles, and 30 for extrema.
We compare the results against a Best Visited (BV) baseline which selects the location the robot took a measurement at that is closest to each quantile value, i.e. solve $\estimatedquantilespatiallocations^* = \argmin_{\estimatedquantilespatiallocations_{bv} \subset \sensedlocations} \|\estimatedquantilevalues - \mu(\estimatedquantilespatiallocations_{bv})\|_2$.

For \rev{SA}, we use $T_\t{max} = 5$, $T_\t{min} = 0.001$, and cooling rate $\t{cr} = 0.005$ which leads to approximately 1000 optimization steps, and reset to the best solution every 100 steps. 
We start the optimization using the solution found by the \rev{BV} baseline.
For \rev{CE}, we use $\alpha = 0.9$, $\eta = 0.9$, 50 samples per iteration, and 100 iterations.
$\alpha$ is a weighting factor on new samples, which is used to prevent premature convergence.
For \rev{BO}, we use the expected improvement acquisition function and initialize the GP with 50 randomly selected $\quantilespatiallocations$s as well as the solution found by the Best Visited baseline.
We \rev{use} 100 iterations and report the best found solution.

\subsubsection{\rev{Discussion}}
The error between the values at the estimated quantile locations and the true quantile values for the drone and the AUV experiments can be seen in \Cref{fig:ecomapper_ps_boxplots}.
Overall, we find that cross-entropy and Bayesian optimization produced the locations with values closest to the true quantile values.
Both these methods perform a global search in the space of possible locations, indicating that global search optimizes \cref{eq:ps_loss} more effectively.
Simulated annealing had greater variability in performance.
We believe this is because it is a local search method and may fail to escape local optima.
Best Visited produced a good initialization for the other methods, but was easily outperformed.
We find that CE performs the best in three out of six possible scenarios.
In particular, we see a 15.7\% mean reduction in median error using CE with our proposed loss function compared to the BV baseline across all environments when using our proposed quantile standard error optimization function for quantile estimation during exploration.

In general, methods can find the best points when selecting for quartiles or deciles,
while the upper extrema are more difficult.
Because the robot is limited in the amount of environment it can explore, the upper extrema  are less likely to be measured during exploration.
This leads to these quantiles being more challenging to select representative points for.

\begin{figure}[t]
    \centering
    \includegraphics[trim={0.5cm .5cm 0.7cm 0.9cm}, clip, width=.45\columnwidth]{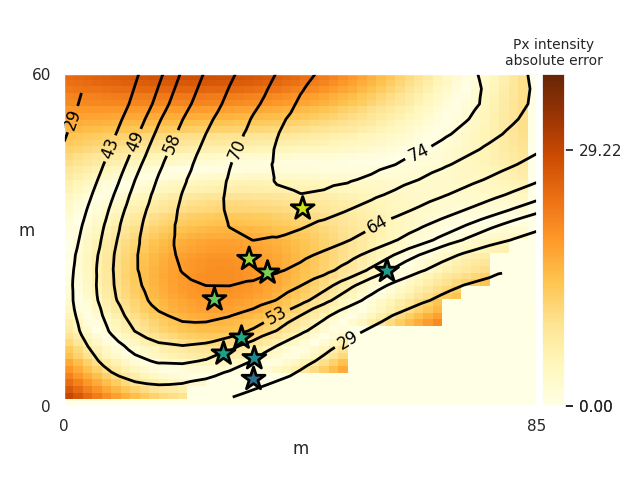}
    %\vspace{-10px}
    \includegraphics[trim={0.5cm .5cm 0.5cm 0.9cm}, clip, width=.45\columnwidth]{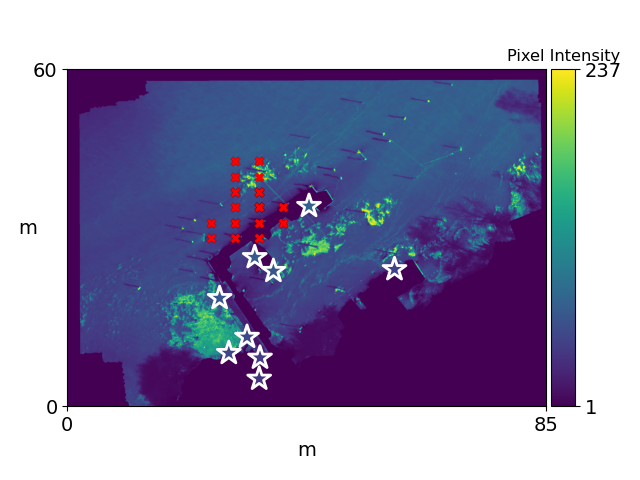}
    
    \caption{
    \textbf{Physical locations (stars) selected by the cross-entropy optimizer for deciles on the drone experiment.}
    \rev{[Left] Black lines: true quantile value contours, overlaid on the absolute error between $\mu(\gtsensedlocations)$ and $GT(\gtsensedlocations)$. Note that the lowest error tends to follow the quantile contour lines.}
    [Right] Red crosses: locations the robot visited, overlaid on the ground truth image.
    }
    \label{fig:contours}
    \vspace{-17px}
\end{figure}

%\textbf{Drone with Camera}
%\Cref{fig:clearlake_ps_boxplots} shows 
%$RMSE(\quantilevalues,\estimatedquantilespatiallocations)$
%for the drone experiments for each of the four point selection methods.

Figure \ref{fig:contours} shows results for one seed of the drone \rev{task} when monitoring deciles.
The suggested locations, shown as stars, align relatively closely with the true quantile values $\quantilevalues$, shown by the contours \rev{on the left image}.
This demonstrates the ability of the optimizers to produce good location suggestions to guide environmental analysis.

\begin{figure}
    \centering
    \includegraphics[trim={1cm .5cm 0.8cm 1.5cm}, clip, width=.65\columnwidth]{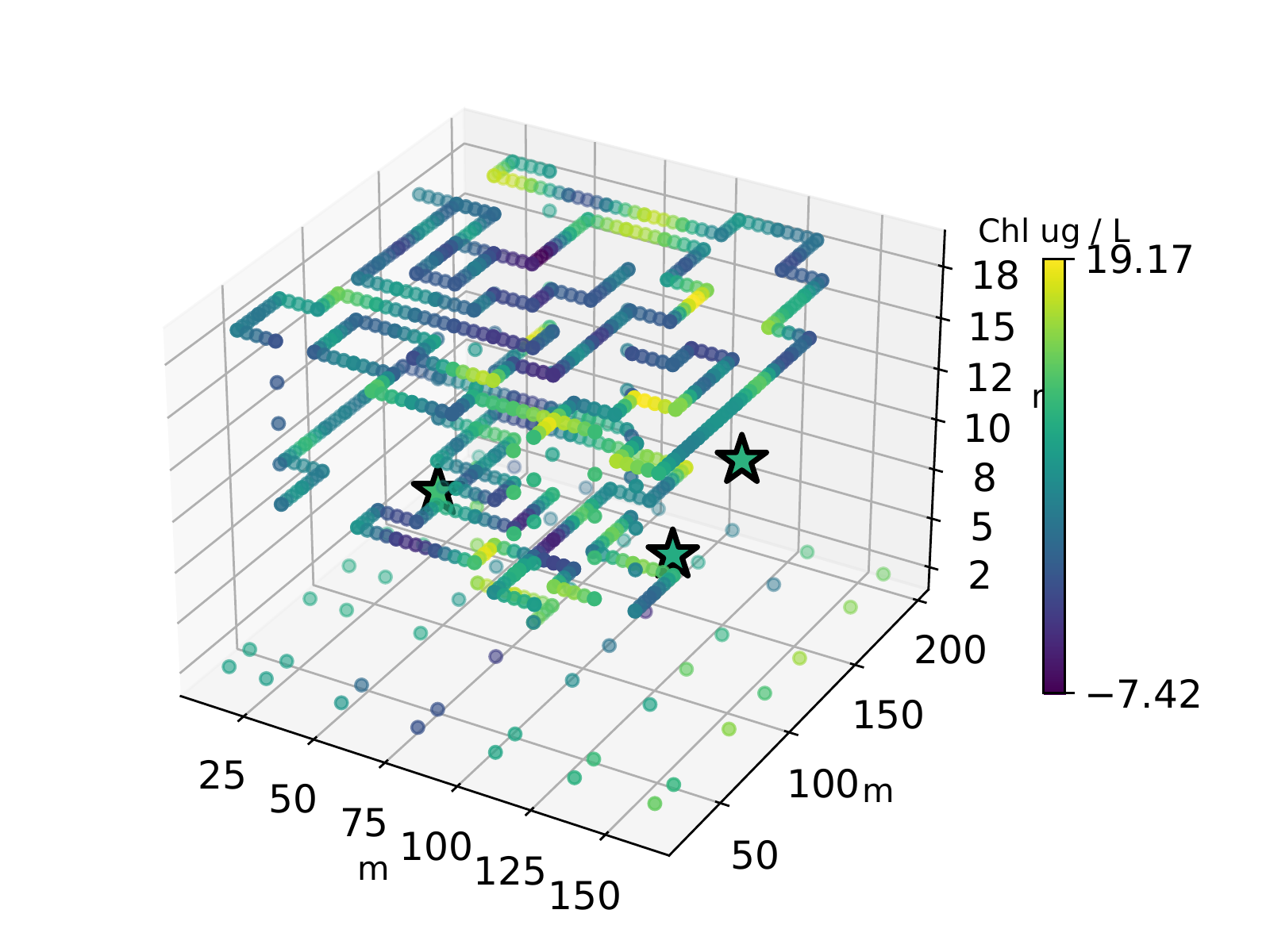}
    \vspace{1px}
    \caption{\textbf{Physical locations (stars) selected by the cross-entropy optimizer for the upper extrema quantiles with an AUV using a chlorophyll point sensor.}
    Blue/green points are the measured locations $\sensedlocations$.
    }
    \label{fig:3d_points}
    \vspace{-17pt}
\end{figure}

The right part of Figure \ref{fig:contours} shows the same locations on top of the orthomosaic of what the drone could measure during exploration.
This part of the figure highlights the difficulty of the problem of IPP for quantiles. 
The robot could only explore 15\% of the total environment.
With only partial knowledge of the distribution, the robot's model of the phenomenon will vary based on the particular points it visited, which in turn affects the estimates of the quantiles.
%
%\textbf{AUV with Chlorophyll Sensor}
% \Cref{fig:ecomapper_ps_boxplots} shows
% $RMSE(\quantilevalues,\estimatedquantilespatiallocations)$
% for the AUV experiments for each of the four point selection methods.
% Like in the drone experiments, the suggested locations correspond well to $\quantilevalues$ produced by the planner.
% Here, cross-entropy performs best for both the extrema and the deciles, demonstrating the effectiveness of global search.
%
\rev{Similarly,} \Cref{fig:3d_points} shows the suggested locations for the upper extrema for one seed \rev{of the AUV task.
For both tasks,} all the optimization methods, with the exception of Best Visited, suggest points that may be spatially far from locations the robot has been able to visit.
This allows for points with values potentially closer to the true quantile values to be selected for scientific analysis.

\begin{figure}[b]
    \vspace{-15px}
    \centering
    \hspace{-10px}
    \includegraphics[trim={3.68cm 2cm 4.18cm 2.3cm}, clip, width=.78\columnwidth, align=c]{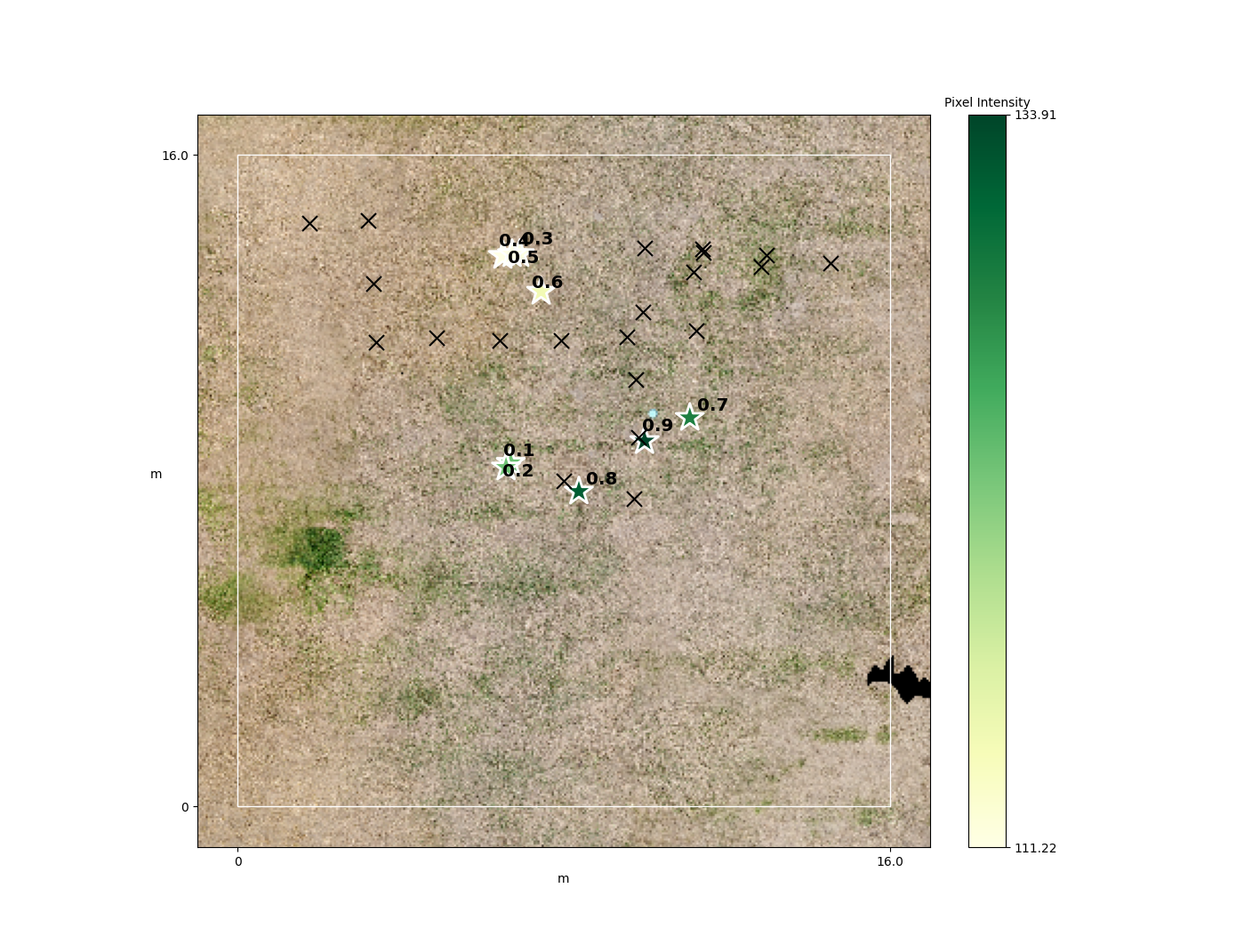}
    %\begin{picture}(0,0)
    %\put(-130,-75){\llap{\includegraphics[trim={80cm 50cm 20cm 30cm}, clip, width=.27\columnwidth]{figures/20210907_144213.jpg}}}
    %\end{picture}
    \hspace{-15px}
    \includegraphics[trim={80cm 0cm 20cm 0cm}, clip, width=.27\columnwidth, align=c]{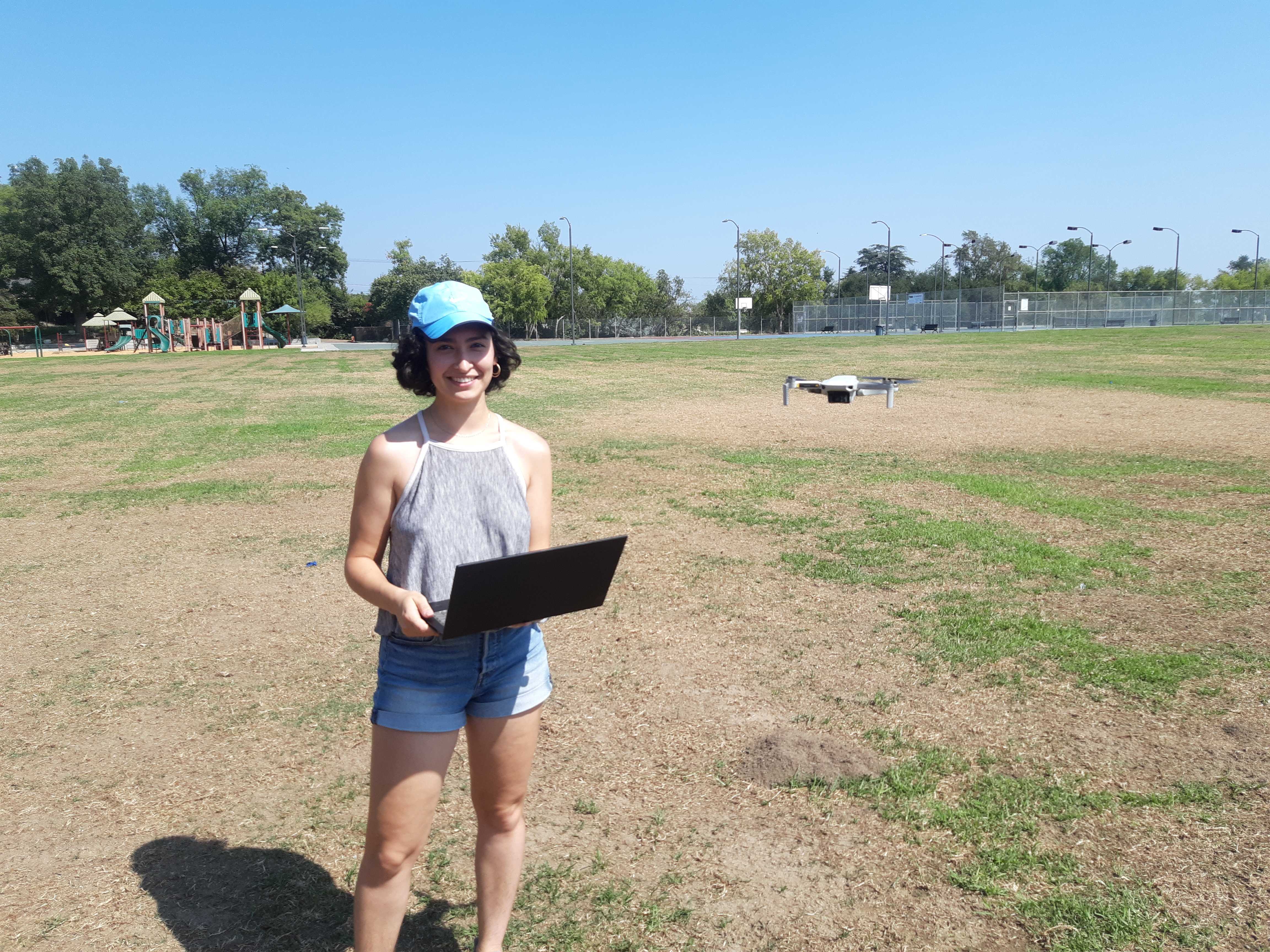}
    %\includegraphics[trim={.18cm .3cm .18cm .3cm}, clip, width=0.9\columnwidth]{figures/field_results.png}
    % \vspace{-5pt}
    \caption{\textbf{Visualization of a field trial modeling a crop health task.}
    \textit{Left:} Crosses are locations where the drone took images. The drone is limited to only visit 20\% of the possible locations to take images.
    The quantiles of interest are the deciles and the locations are chosen by cross-entropy.
    The 9 stars show locations suggested to collect physical specimen. 
    The measurement of interest is the amount of green in each pixel. 
    \rev{\textit{Right:} The drone used in the field trial, in flight.}
    }
    \label{fig:field-results}
    \vspace{-15px}
\end{figure}

\subsection{Field trial}\label{sec:field_trial}
\rev{The goal of our final experiment is to demonstrate our method on real hardware} for a crop health monitoring task \rev{in an open, grassy field}, where the objective is to estimate the deciles of the green present in each pixel of the images (we use green as a proxy for plant health).

\subsubsection{\rev{Setup}}
We use a commercial, off-the-the shelf drone with a standard camera to take measurements of the field at a constant height of 3m. 
Similar to the simulated drone, the drone in this experiment moves in a 2D plane with a north-fixed yaw.
The drone \rev{has a limited budget of} 20 pictures (planning steps) in a $16\times16$ m square grid, where $|\gtlocations| = 10\times10$ locations, and each picture is downsampled to $8 \times 5$ px.
For planning, we use the quantile change objective function because it performs well for camera sensors and is faster than quantile standard error.

\subsubsection{\rev{Discussion}}
Figure \ref{fig:field-results} shows the resulting suggested locations using CE based on the 20 steps the robot took. 
We find that, although the robot cannot explore the entire workspace due to \rev{its limited budget}, and in fact does not visit a large green area, the system is able to suggest varied locations for specimen collection.
\rev{Some selected locations (e.g., those representing the $0.3, 0.4, 0.5$ quantiles) are spatially close to each other, which suggests this area contains a large gradient, or those quantile values are similar.
As the goal is to suggest locations that contain the estimated quantile values, such locations may be near one another.}

\section{Conclusion}
% \imr{could cut some of the background/motivation here if we need space}
Scientists traditionally collect physical specimens at locations selected using heuristics.
They later analyze these specimens in a laboratory to characterize a phenomenon of interest (e.g., the distribution of algae in the water).
We propose to, instead, choose these specimen collection locations by first performing an informative path planning survey with a robot and then proposing locations which correspond to the quantiles of interest of the distribution.

To accomplish this, we propose two novel objective functions, quantile change and quantile standard error, for improving the estimates of quantile values through informative path planning. 
We test these in three settings: a drone with a camera sensor over lake imagery, an underwater vehicle taking chlorophyll measurements, and a field trial using a drone for a crop health task.
Our objective functions perform well and outperform information-theoretic and Bayesian optimization baselines. 
In our experiments, our proposed quantile standard error objective function has a 10.2\% mean reduction in median error when compared to entropy.
%as the objective function. 
% We show that quantile change works well on a task where a drone takes samples with a camera and that quantile standard error performs well on both tasks. 

We also show that \rev{using our proposed loss function with} black-box optimization methods can select environment locations for analysis that are representative of a set of quantiles of interests. 
We find that a cross-entropy optimizer using our loss function outperforms a baseline of using the best measured points,
with a 15.7\% mean reduction in median error in values across all environments.

Our approach can be used to guide physical specimen collection in real field scenarios, as demonstrated by our field trial.
\rev{In recent work, we have discovered that some computational cost can be alleviated with a randomized sampling approach when evaluating the GP in the objective functions.
We believe combining that and an approximate GP can improve scalability.}
\rev{Additionally, experimentally finding parameters was difficult for both the baselines and our proposed methodology, particularly for the GP; this is a direction for future work.}
We also plan to incorporate our specimen collection location suggestions into a larger field campaign involving multiple scientists.

%\includegraphics[scale=1.5]{figures/image (21).png}
%
% ---- Bibliography ----
%
%\begin{thebibliography}{1}
%
%\clearpage
\bibliographystyle{IEEEtran}
\bibliography{references}

%\end{thebibliography}

\end{document}